\documentclass[10pt]{article} 
\usepackage[preprint]{tmlr}

\usepackage{amsmath,amsfonts,bm}

\def\eqref#1{equation~\ref{#1}}

\def\1{\bm{1}}

\DeclareMathAlphabet{\mathsfit}{\encodingdefault}{\sfdefault}{m}{sl}
\SetMathAlphabet{\mathsfit}{bold}{\encodingdefault}{\sfdefault}{bx}{n}

\usepackage{xcolor}
\usepackage{graphicx}
\usepackage{algorithm}
\usepackage{algpseudocode}
\usepackage{bm}
\usepackage{bbm}
\usepackage{amsmath}
\usepackage{amssymb}
\usepackage{amsthm}
\usepackage{multirow}
\usepackage{subcaption}
\usepackage{comment}

\newtheorem{example}{Example}
\newcommand{\indep}{\perp \!\!\! \perp}
\newcommand\numberthis{\addtocounter{equation}{1}\tag{\theequation}}
\newcommand{\update}[1]{\color{blue}}

\title{Active \& Passive Causal Inference: Introduction}

\author{\name Daniel Jiwoong Im \& Kyunghyun Cho \email \{ji641,kc119\}@nyu.edu \\
      \addr Center for Data Science\\
      New York University}

\def\month{MM}  
\def\year{YYYY} 

\begin{document}

\maketitle

\begin{abstract}
This paper serves as a starting point for machine learning researchers, engineers and students who are interested in but not yet familiar with causal inference. We start by laying out an important set of assumptions that are collectively needed for causal identification, such as exchangeability, positivity, consistency and the absence of interference. From these assumptions, we build out a set of important causal inference techniques, which we do so by categorizing them into two buckets; active and passive approaches. We describe and discuss randomized controlled trials and bandit-based approaches from the active category. We then describe classical approaches, such as matching and inverse probability weighting, in the passive category, followed by more recent deep learning based algorithms. By finishing the paper with some of the missing aspects of causal inference from this paper, such as collider biases, we expect this paper to provide readers with a diverse set of starting points for further reading and research in causal inference and discovery.
\end{abstract}

\section{Introduction}


The human curiosity and the desire to understand how things work lead to studying causality \citep{Celeste2015, Zheng2020HowCI, ferraro2019, Bender2020}. Causality is about discovering the causal relationship between variables. Causality delineates an asymmetric relationship where it can only say {\em “A causes B”} or {\em “B causes A”}, while correlation expresses a symmetric relationship that measures the co-occurrence of A and B. They can extend to more than two variables. Being able to depict the causal relationship is an ideal framework for humans to explain how the system works. 

An important concept in causality, that we are particularly interested in, is causal effect. It refers to the impact of a choice of action on an outcome. For example, what is the impact of receiving COVID-19 vaccine on catching COVID-19?  In order to know the effect of Covid-19 vaccination, we must be able to predict the outcomes of both taking and not taking the vaccine, respectively. That is, we must know the potential outcome of taking an arbitrary action. We call this process of inferring the potential outcome given an action causal inference (CI) \citep{rubin1974estimating}.

Correlation does not imply causation, and at the same time, it is possible to have causation with no correlation. CI is tricky like that because observation is not the same as intervention.
The former is about passively observing what happens, while the latter is about observing what happens when one actively intervenes in the world.
Taking the CI approach allows us to distinguish between correlation and causation by clearly defining the two probabilities.

To infer causal effects we must measure intervention probabilities. Intervention probability is the probability of a particular outcome resulting from our intervention in the system by imposing a specific action. This is different from conditioning, as we actively alter the action. We often have access to conditional and joint probabilities, but not intervention probabilities directly. It has thus been a major research topic to infer the intervention probability from conditional and joint probabilities in many disciplines \citep{rubin1974estimating, berry1985bandit, Heckman1997, Weitzen2004, Hernan2006, Breslow2009, Graepel2010, Banerjee2012, Yazdani2015, Bouneffouf2020}.\\

There are two main frameworks to introducing estimating causal effect, \citet{rubin1974estimating}'s potential outcome framework and \citet{Pearl09a}'s do-calculus framework. Potential outcome framework focuses on the concept of the outcomes that would have been observed under different treatment conditions. Do-calculus revolves around a set of formal rules for reasoning about intervention in a causal model.  This introductory paper diverges from conventional teaching methods in causal inference by combining both Rubin’s framework of potential outcomes and Judea Pearl’s framework of do-calculus. We adopt a holistic approach, drawing upon concepts from both paradigms to construct a foundational understanding of causal inference rooted in first principles.
Our choice to introduce potential outcomes initially stems from its intuitive appeal, particularly when illustrating treatment effects through familiar examples from domains such as medicine or the sciences. However, as we delve deeper into the formalization of causal models, the incorporation of intervention probabilities becomes essential, necessitating a shift towards joint and conditional probability distributions. By incorporating aspects of both frameworks, we aim to present a unified perspective on causal inference that facilitates a smoother transition between the intuitive conceptualization of potential outcomes and the more formalized treatment of intervention probabilities.

\begin{figure}[t]
        \centering
        \includegraphics[width=0.8\linewidth]{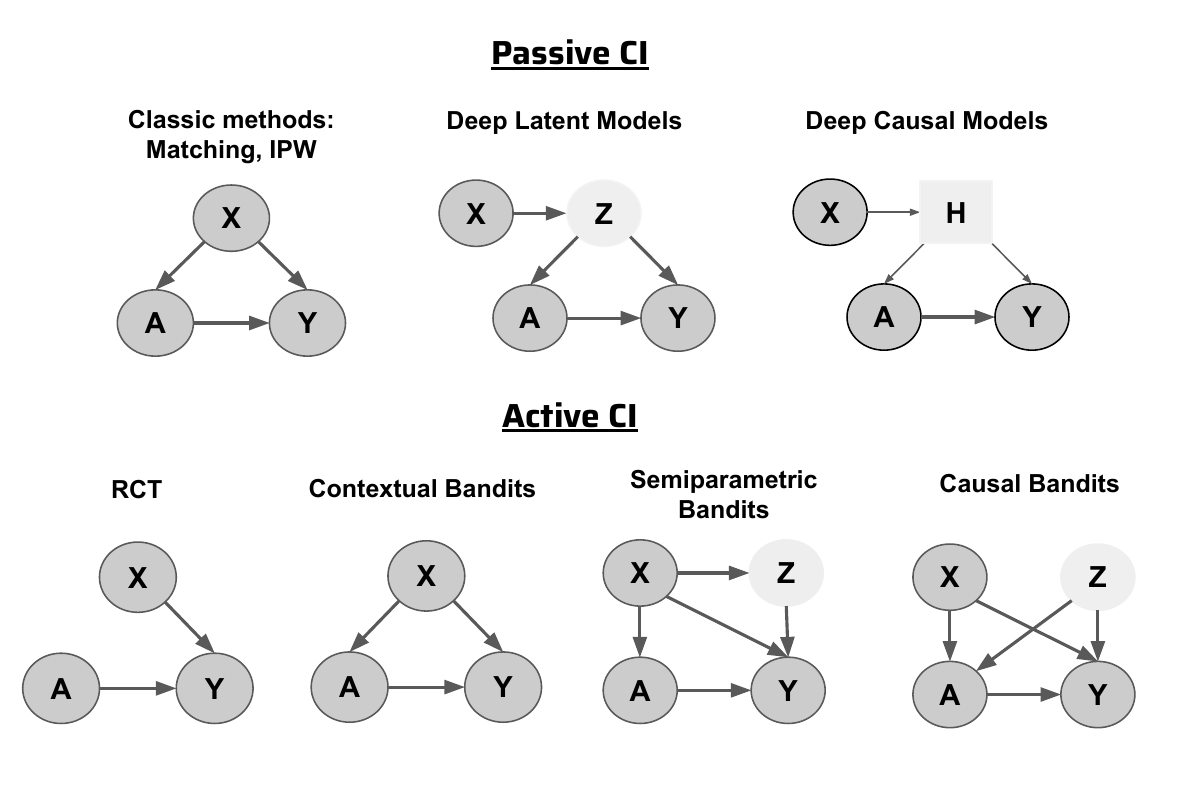}
        \caption{Examples of passive and active causal inference methods. Dark gray nodes correspond to observed variables while light gray nodes correspond to latent variables. A square node corresponds to a deterministic variable while a circle corresponds to stochastic variables.}
        \label{fig:ci_causal_graph}
\end{figure}

A causal graph is a graphical representation of causal relationships among variables in a system. It visually depicts how variables influence each other, helping us to understand and analyze the causal structure of a phenomenon.
Figure~\ref{fig:ci_causal_graph} shows the graph representation of causal relationships for a variety of CI methods.
Depending on the data collection process and experimental setup, certain methods don't require knowing the causal graph in priority, such as RCT and difference-in-difference method, while other methods require knowing the structure of a causal graph.
In this paper,
we consider a problem setup in which 
we have a covariate $X$, an action $A$, and an outcome $Y$.
The action $A$ is a variable of interest and has a causal effect on the outcome. 
The outcome $Y$ is a variable that is affected by the treatment and is what we often want to maximize.
Covariates $X$ are all the other variables that may affect and be affected by the action $A$ and the outcome $Y$. We are particularly interested in the case where these covariates are confounders, i.e., affect the action and the outcome together.
We measure how treatments affect outcomes by looking at both the average effect across groups and how each person's treatment affects them personally.

There is enormous work on various assumptions and conditions that allow us to infer causal effects \citep{rubin1974estimating}. The most fundamental assumptions are i) exchangeability, ii) positivity, iii) consistency, and iv) no interference. 
These assumptions must be satisfied at the time of data collection rather than at the time of causal inference. When these assumptions are met, we can then convert statistical quantities, estimated from collected data, into causal quantities, including the causal effect of the action on the outcome
\citep{Miguel2019, Rashelle2019, Zheng2019}.  One way to satisfy all of these assumptions is to collect data actively by observing the outcome after randomly assigning an action independent of the covariate. Such an approach, which is often referred to as a random controlled trial (RCT), is used in 
clinical trials, where patients are assigned randomly to an actual treatment or placebo \citep{Chalmers1981, Kendall12003, Aronow2021}. RCT is deliberately designed to prevent confounding the treatment and the outcome, 
so that 
the conditional probabilities estimated from collected data approximate the intervention probabilities as well. 

Randomized data collection is not always feasible and often suffers in efficiency from running large-scale experiments. There has been an enormous amount of work from various disciplines on estimating causal effects without such randomized data collection~\citep{rubin1977assignment, Rubin1979UsingMM, Chalmers1981, Lu2004OptimalPM}. As an alternative, different approaches have been proposed, including figuring out how to work with the non-randomized dataset and finding a more efficient way to collect data than the randomized approach. In this paper, we organize these CI methods into passive and active learning categories. In the passive CI category exist methods that work {\it given} a dataset which was {\it passively} collected by the experts. In contrast, the active CI category includes methods that may actively intervene in the data collection process. RCT for instance belongs to the active CI category, as it actively collects data by randomizing the treatment assignment. There are however other methods in the same category that aim also to maximize the outcome by making
a trade-off between exploration and exploitation.  

The organization of this literature review paper is as follows. In \S2, we introduce the definitions and metrics for estimating causal effects and discuss in depth the assumptions necessary for identification of causal effects. We then cover naive conditional mean estimator and ordinary square estimator, both of which are widely used with randomized datasets \citep{rubin1974estimating, Pearl2010}. In this paper, We do not consider collider bias and we assume a stationary conditional probability distribution.

In \S3, we describe RCT and move on to bandit approaches in the active CI category. While bandits are used in many practical applications, the research community has been adding more emphasis on theoretical analysis of minimizing the regret bounds of different policy algorithms \citep{berry1985bandit, Langford2007}. We look at bandits through the lens of CI, where many of the bandit algorithms can be seen as learning the classic causal graph in Figure~\ref{fig:condition_intervention_graph} with different exploration and exploitation rates. We examine different constrained contextual bandit problems that correspond to different causal graphs, respectively. We also compare passive CI learning methods to bandits on naive causal graphs. We furthermore review causal bandits which consider graphs with unknown confounding variables \citep{Bareinboim2015, Lattimore2016, Sachidananda2017OnlineLF}. In this survey, we limit our scope to bandits and do not consider causal reinforcement learning which we leave for the future.

In \S4, we start with classical approaches in the passive CI category, such as matching \citep{RUBIN1992, Gu1993ComparisonOM}, inverse probability weighting \citep{ROSENBAUM1983, RUBIN1992, Hirano2003} and doubly robustness methods \citep{Bang2005, Shardell2014, Seaman2018}. 
We then discuss deep learning based CI methods~\citep{Cai2012, Pearl2015, Johansson2016, Wang2016,Louizos2017}. 
Deep learning is particularly useful when we need to conduct causal inference on 
high dimensional data with a very complicated mapping from input to output, as deep neural networks can 
learn a compact representation of action as well as covariate that captures the intrinsic and semantic similarities underlying the data \citep{Kingma2014, Rezende2014}. 
Deep learning is applied to CI in order to infer causal effects by learning the hidden/unknown confounder representations from complicated data and causal graph relationships. 
Such a capability of learning a compact representation from a high-dimensional input allows it to 
work with challenging problems such as those involving raw medical images and complex treatments \citep{Castro2020, Im2021, Puli2022, Amsterdam2022}.

CI is an important topic in various disciplines, including statistics, epidemiology, economics, and social sciences, and is receiving an increasingly higher level of interest from machine learning and natural language processing due to the recent advances and interest in large-scale language models and more generally generative artificial intelligence. 
In this paper, we cover various CI algorithms and categorize them into active and passive CI families.
The goal of this paper is to serve as a concise and readily-available resource for those who are just starting to grow their interest in causal inference.

\newpage
\section{Background}
\subsection{Preliminary}
\label{sec:preliminary}

\begin{figure}[t]
    \begin{minipage}{0.5\linewidth}
        \centering
        \includegraphics[width=\linewidth]{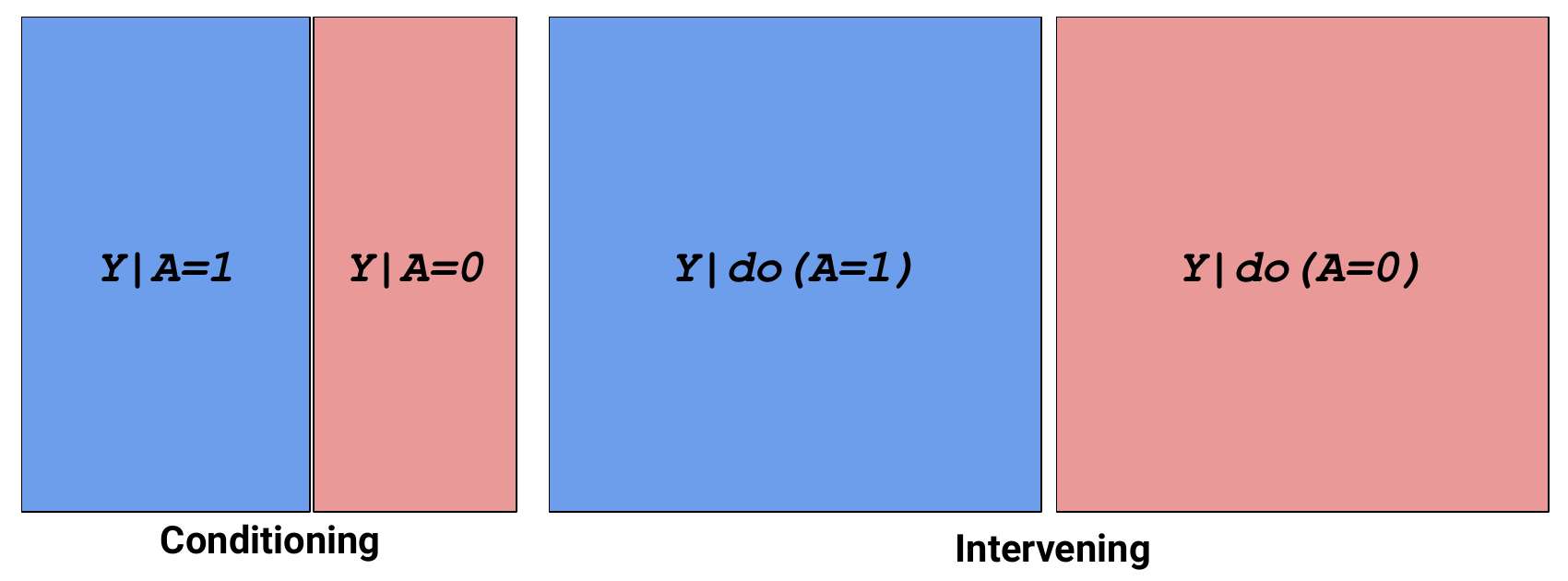}
        \subcaption{Population}
        \label{fig:condition_intervention_pop}
    \end{minipage}
    \begin{minipage}{0.5\linewidth}
        \centering
        \includegraphics[width=\linewidth]{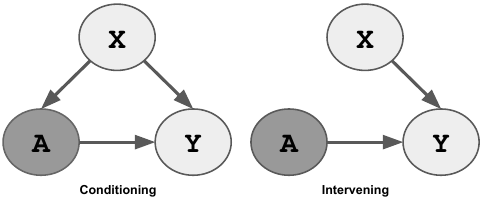}
        \subcaption{Causal Graph}
        \label{fig:condition_intervention_graph}
    \end{minipage}
    \caption{Condition versus intervention}
    \label{fig:condition_intervention}
\end{figure}

Let $X$, $A$ and $Y$ be {\it covariate}, {\it action}, and {\it outcome} variables, respectively. 
We define the {\it joint}, {\it conditional} and {\it intervention} probabilities as
\begin{align*}
    \text{Joint: } & p(Y=y,A=a,X=x)=p(X=x)p(A=a|X=x)p(Y=y|A=a,X=x), \\
    \text{Conditional: } & p(Y=y|A=a)=
    \frac{\sum_x p(X=x)p(A=a|X=x)p(Y=y|A=a,X=x)}
    {\sum_{x',y'} p(X=x')p(A=a|X=x')p(Y=y'|A=a,X=x')}, \text{and}\\
    \text{Intervention: } & p(Y=y|do(A=a))=\sum_x p(X=x)p(Y=y|A=a,X=x),
\end{align*}
respectively.
We marginalize the covariate variable to obtain 
the conditional and intervention probabilities. 
We observe that the conditional probability is often different from the intervention probability, $p(Y|A=a) \neq p(Y|do(A=a))$. 
The conditional probability takes into account the prevalence of a particular action $a$ in the population and checks how often a particular outcome $y$ is associated with it. 
On the other hand, the intervention probability does not consider the prevalence of the action $a$ and only considers what the outcome would be had the action been forced to be $a$. 
$p(Y|A=a,x)$ distribution tells us the effect of $A$ on $Y$  given $X=x$, since $A=a$ is assigned directly.  $p(Y|do(A=a))$ is a marginal distribution of $p(Y|A=a,x)$ over $X$ and $A=a$ was directly assigned. As soon as one incorporates $p(A=a|X)$ or $p(A=a)$, it is not an intervention but a joint.
See Figure~\ref{fig:condition_intervention_pop} for a graphical illustration.
This can be understood as removing all the incoming edges to the action variable, i.e. $X \to A$, when we intervene on $A$, as in Figure~\ref{fig:condition_intervention_graph}.
The intervention probability formula above reflects it by ignoring the probability of a particular action. 
The intervention probability $p(Y|do(A))$ describes the causal effect of action $A$. The corresponding causal graph $A\Rightarrow Y$ is a graphical representation used in causal inference to depict the causal relationships between variables in a system (see Figure~\ref{fig:condition_intervention_graph} right).

Consider the following example that shows how intervention and conditional probabilities are sometimes different and sometimes the same, given two variables.
\begin{example}
    let $C$ indicate whether coffee is hot or cold and $T$ be the thermometer reading that measures the temperature of the coffee. 
    Based on our everyday observation, $T$ and $C$ are highly correlated. 
    For instance, $p(C=\text{hot}|T=70^{\circ})$ and $p(T=70^{\circ} | C=\text{hot})$ are both high. 
    However, it is clear that forcing the thermostat's reading to be high does not cause the temperature of coffee to go up, that is, $p(C=\text{hot}|do(T=70^{\circ}))$ is low despite high $p(C=\text{hot}|T=70^{\circ})$. 
    On the other hand, boiling coffee would indeed cause the thermostat's reading to go up, that is, both $p(T=70^{\circ} | C=\text{hot})$ and $p(T=70^{\circ} | do(C=\text{hot}))$ are high.
\end{example}

Another example, demonstrating the discrepancy between the intervention and conditional probabilities, is {\it Simpson's paradox} in which different assumptions about outcome, treatment, and confounder variables lead to different conclusions on causation: 
    \begin{example}[Simpson's paradox illustration \citep{Carlson2019}]\label{ex:simpson_paradox}
        You are studying sex bias in graduate school admission. 
        According to the data, men were more likely to be admitted to graduate school than women were, where 40\% of male applicants and 25\% of female applicants were admitted. 
        In other words, there was a strong association between being a man and being admitted. 
        You however found that such association varies across different sub-populations.
        For example, 80\% and 46\% of men and women were admitted to natural science respectively, and 
        20\% and 4\% of women and men were admitted to social science respectively. 
        It turns out that the social science department has a much lower acceptance rate than the natural science department, while
        women were more likely to apply to social science and men were more likely to apply to natural science departments.
        In summary, we can derive different conclusions about sex and admission rate association by either combining or separating sub-populations.
        
        We can notice how adding the department to the covariate variable or not changes the result.
        If the department is not part of the covariate, then the conditional and intervention probability coincide with each other, since there is no confounding via the department choice. Otherwise, these two probabilities deviate from each other due to $p(S=s|X=x)$. This demonstrates that CI is inherently dependent upon our modelling assumption.
    \end{example}

A {\it potential outcome} is an outcome given a covariate under a potential action  \citep{rubin1974estimating,rubin2005causal}.
Assuming a binary outcome variable $Y\in \lbrace 0,1 \rbrace$ and a discrete action variable $A\in \mathcal{A}$, the potential outcome \citep{rubin1974estimating} is defined as
\begin{align*}
    Y_X(a) = Y|do(A=a).
\end{align*}
For instance, there are two potential outcomes, $Y_X(1)$ and $Y_X(0)$, for a binary action $A\in \lbrace 0, 1\rbrace$.
It is often impossible to compute these potential outcomes given a fixed covariate directly due to the fundamental problem of causal inference, 
that is, we cannot perform two distinct actions for a given covariate $X$ simultaneously and observe their outcomes~\citep{Saito2019}.
We can only observe the outcome of one action that has been performed, to which we refer as the {\it factual outcome}, but cannot observe that of the other action, to which we refer as the {\it counterfactual outcome}. 
We instead consider the potential outcome averaged over the entire population, represented by the covariate distribution $p(X)$. We call it the {\it expected potential outcome} 
$\mathbb{E}_{Y,X}[Y_X(A=a)]$, as opposed to the {\it conditional potential outcome},
where we marginalize out the covariate $X$. 

The goal of causal inference is to estimate the potential outcomes that each individual would have experienced under different treatment conditions, as well as the average or expected outcomes across the population, based on observed data,
$(y_{i}|do(A=a_{i}),x_{i},a_{i}) \overset{\mathrm{iid}}{\sim} p$, where $p$ is the data distribution and $a_{i}\in \mathcal{A}$ is the action performed for the $i$-th covariate $x_i$. 

We often hear about the {\it treatment effect} of an experimental drug or a new surgical procedure in the medical literature.
Treatment effect measures whether and how much the treatment caused the difference in the outcome (or in the potential outcome). 
Often treatment effect is used for binary actions in medical research.
Unless confusing, we will interchangingly use the treatment effect and causal effect throughout this paper.
In general, the treatment
effect is defined as the difference between two potential outcomes $Y_X(1) - Y_X(0)$.
The {\it average treatment effect} (ATE) is then the difference between the potential outcomes averaged over the covariate distribution \citep{rubin1974estimating, imbens2004nonparametric}:
\begin{align}
    ATE := \mathbb{E}_{X,Y}[Y_X(1) - Y_X(0)] = \mathbb{E}_{X,Y}[Y_X(1)] - \mathbb{E}_{X,Y}[Y_X(0)].
\end{align} 

We may be interested in the average treatment effect over a subpopulation, defined by a subset $X' \subseteq X$ of covariates. 
We then compute the {\it conditional average treatment effect} (CATE).
CATE is defined as averaging the treatment effect for an individual patient characterized by $X'$ \citep{Radcliffe2007UsingCG, Athey2015}:
\begin{align}
    CATE(x'):= \mathbb{E}_{Y, X \backslash X'} [Y_X(1) - Y_X(0)|X'=x'] = \mathbb{E}_{Y, X \backslash X'}[Y_X(1)|X'=x'] - \mathbb{E}_{Y, X \backslash X'}[Y_X(0)|X'=x'],
    \label{eqn:cate_bianry}
\end{align}
where $X \backslash X'$ is the remainder of the covariate over which we compute the expectation. 

There are a few alternatives to the ATE. 
The first one is an {\it expected precision in the estimation of 
heterogeneous effect} (PEHE \citep{imbens2004nonparametric}):
\begin{align*}
    PEHE:= \mathbb{E}_{X,Y}[(Y_X(1) - Y_X(0))^2].
\end{align*}
Another alternative is 
the {\it population average treatment effect} (PATT) for the treated, i.e. $a=1$ \citep{rubin1977assignment, HECKMAN1985239}:
\begin{align*}
    PATT(a):= \mathbb{E}_{X,Y | A=a}[Y_X(1) - Y_X(0)],
\end{align*}
which is a helpful quantity when a particular, treated sub-population is more relevant in the context of narrowly targeted experiments. 
All of these metrics have their own places. 
For instance, it is more common to see PEHE in medical research, while PATT can be used to study the effect on the treated group programs (e.g. individuals disadvantaged in the labour market~\citep{HECKMAN1985239}).

\subsection{Assumptions for Causal Inference}
\label{sec:assumptions}

Unfortunately, we cannot simply average the factual outcomes for each action to estimate the average treatment effect, because this corresponds to measuring the expected conditional outcome which is
a biased estimate of the expected potential outcome. 
This bias is usually due to confounders.
For instance, socioeconomic status can be a confounder in the study of the treatment effect of a medication. 
Socioeconomic status often affects both patients' access to medication and their general health, which makes it a confounder between the treatment and health outcome. 
In this case, the expected conditional outcome estimate is biased, because those patients who receive the medication are also likely to receive better healthcare, resulting in a better outcome. 
We must isolate the effect of medications on the health outcome by separating out the effect of having better access to healthcare due to patients' higher socioeconomic status, in order to properly estimate the expected treatment effect. 
In this section, we review the assumptions required for us to obtain an unbiased estimator for the average potential outcome.

The main strategy for estimating the (average) potential outcome is to compute causal quantities, such as intervention probabilities, from statistical quantities that are readily estimated from a set of samples, i.e., a dataset. 
In this context, we can say that a causal quantity is {\em identifiable} if we can compute it from statistical quantities. In doing so, there are a number of assumptions that must be satisfied.

\paragraph{Positivity/Overlap.}

The first step in estimating the potential outcome is to estimate the conditional probabilities from data. In particular, we need to compute 
\begin{align*}
    p(Y|X,A) = \frac{p(Y, A|X)}{p(A|X)}
\end{align*}
for $X$ with $p(X) > 0$. This implies that $p(A=a|X)$ for the action $a$, of which we are interested in computing the potential outcome, must be positive. We call it the {\it positivity}.  Positivity is a necessary condition for computing ATE as we need the $p(A|X=x) >0 $ in the denominator for data $x$.
%
The {\it overlap} assumption is similar to the positivity assumption but applies to the covariate. It requires that 
the distributions $p(X|A=0)$ and $p(X|A=1)$ have common support. 
Partial overlap occurs when you are missing on particular action for a certain area of covariate space. For example,
we can have treated units of certain patient groups but no control units, or vice versa.

\paragraph{Ignorability/Exchangeability.}

Even if we can estimate the statistical quantities, such as the conditional probability $p(Y|X,A)$, we need an additional set of assumptions in order to turn them into causal quantities. The first such assumption is {\it exchangeability} which states that the potential outcome $\hat{Y}(a)$ must be preserved even if the choice of an action to each covariate configuration $p(A|X)$ changes. That is, the causal effect of $A$ on $Y$ does not depend on how we assign an action $A$ to each sample $X$. This is also called {\it ignorability}, as this is equivalent to ignoring the associated covariate when assigning an action to a sample, i.e., $A \indep X$. This enables us to turn the conditional probabilities into intervention probabilities, eventually allowing us to estimate the potential outcome, which we describe in more detail later. 

The exchangeability is a strict condition that may not be easily met in practice. This can be due to the existence of confounding variables, selection bias, or due to time-dependency between action selections (violation of Markov Assumption).  We can relax this by assuming {\it conditional exchangeability}. As we did for defining the CATE above, we partition the covariate into $X$ and $X'$ and condition the latter on a particular value, i.e., $X'=x'$. If the exchangeability is satisfied conditioned on $X'=x'$, we say that conditional exchangeability was satisfied. This however implies that we are only able to estimate the potential outcome given a particular configuration of $X'$. To meet the ignorability assumption, we need to achieve unconfoundedness, which refers to having an absence of confounding variables in a causal relationship. This involves careful design, data collection, and measurements to control potential confounders and isolate the impact. These two assumptions together allow us to turn the statistical quantities into causal quantities, just like when (unconditional) exchangeability alone was satisfied. 

\paragraph{Consistency and Markov assumption.}

Data for causal inference is often collected in a series of batches rather than at once in parallel. Each $t$-th batch has an associated potential outcome $\hat{Y}(a_t)$ for action $a$. If the potential outcome changes over these batches, we cannot concatenate all these batches and use them as one dataset to estimate the potential outcome. That is, we must ensure that $\hat{Y}(a_t) = \hat{Y}(a_{t'})$ for all $t, t'$ where $a_t = a_{t'}$. This condition is called {\it consistency}. Furthermore, we must ensure that the past batches do not affect the future batches in terms of the potential outcome, i.e., $\hat{Y}(a_1, \ldots, a_t) = \hat{Y}(a_t)$, as this would effectively increase the action space dramatically and make it impossible to satisfy positivity. We call this condition a {\it Markov} assumption. 

In practice, the potential outcomes are estimated from the data, which is a statistical estimation. Hence, all assumptions are required to turn causal estimand into statistical estimand. We show step-by-step how each assumption is used to compute ATE:
    \begin{align*}
        ATE &= \mathbb{E}_X [\mathbb{E}_{X'}[Y(1) - Y(0)|X']] \\
            &= \mathbb{E}_X [\mathbb{E}_{X'}[Y(1)|X'] - \mathbb{E}_{X'}[Y(0)|X']] \tag*{\text{(Conditional exchangeability)}}\\
            &= \mathbb{E}_X [\mathbb{E}_{X'}[Y(1)|A=1,X']] -  \mathbb{E}_X [\mathbb{E}_{X'}[Y(0)|A=0,X']] \tag*{\text{(Ignorability)}}\\
            &= \mathbb{E}_X [\mathbb{E}[Y|A=1,X]] -  \mathbb{E}_X [\mathbb{E}[Y|A=0,X]]\tag*{\text{(Consistency)}}\\
            &= \mathbb{E}[Y|A=1] -  \mathbb{E}[Y|A=0]].
    \end{align*}

\subsection{Discussion on the assumptions in practice}
\label{sec:assumption_discussion}

These assumptions above, or some of their combinations, 
enable us to derive causal quantities from statistical quantities. 
It is thus important to carefully consider these assumptions when faced with causal inference and how they may be violated, as most of these are often impossible to verify in practice. 
We discuss a few important points regarding these assumptions in practice.

\paragraph{Unconfoundedness \& Conditional Exchangeability.}

We face the challenge of verifying whether the potential outcome remains the same across all possible confounder configurations because we cannot enumerate all possible confounders.
In practice, we use conditional exchangeability because we often condition each individual data point (e.g., covariates per patient).
However, the conditional exchangeability assumption is impossible to test and verify in practice.
Estimating the potential outcome given a particular configuration of $X'$ means removing all the existing confounders for $X'$, however, we cannot know every possible confounder out there. 
In practice, we condition $X'=x'$ to be an individual data point (e.g., $x'$ being a patient) and often conditional exchangeability is taken for granted for that data point. 
    
\paragraph{Unconfoundedness vs. Overlap.}

In order to estimate ATE, one must assume both unconfoundedness and positivity. 
It is however difficult to satisfy both of them in practice, because there is a natural trade-off between them. We are likely to satisfy the unconfoundedness by adding more features to the covariate, which in turn increases the dimensionality of data. This in turn increases the chance of violating the overlap assumption due to the curse of dimensionality. Similarly, we can satisfy the overlap assumption by 
choosing only the minimum number of features as a covariate, 
but we may unintentionally create unobserved confounders along the way.\\

\begin{figure}[t]
    \begin{minipage}{0.5\linewidth}
        \centering
        \includegraphics[width=\linewidth]{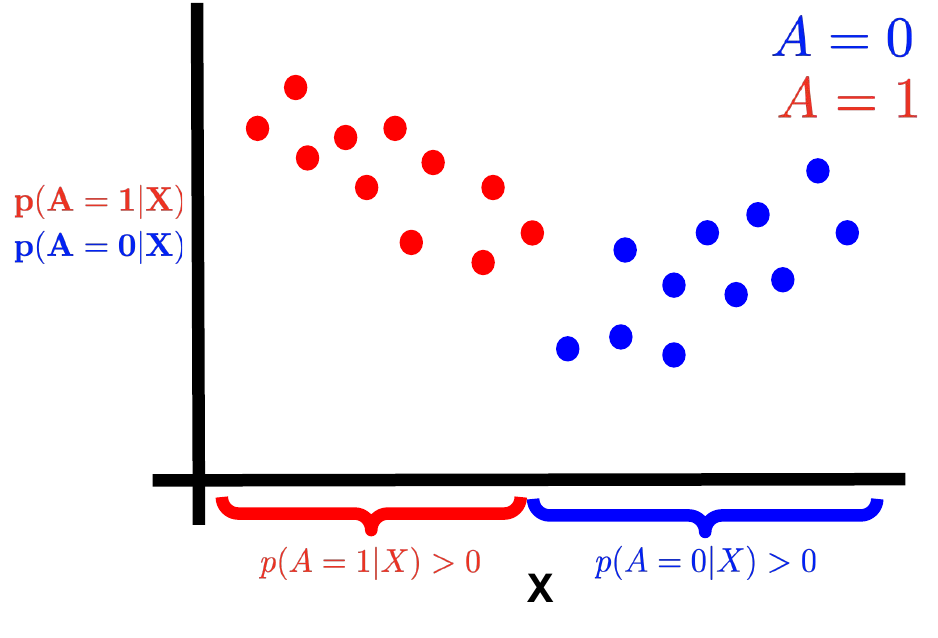}
        \subcaption{Generalization through extrapolation}
        \label{fig:extrapolation}
    \end{minipage}
    \begin{minipage}{0.5\linewidth}
        \centering
        \includegraphics[width=\linewidth]{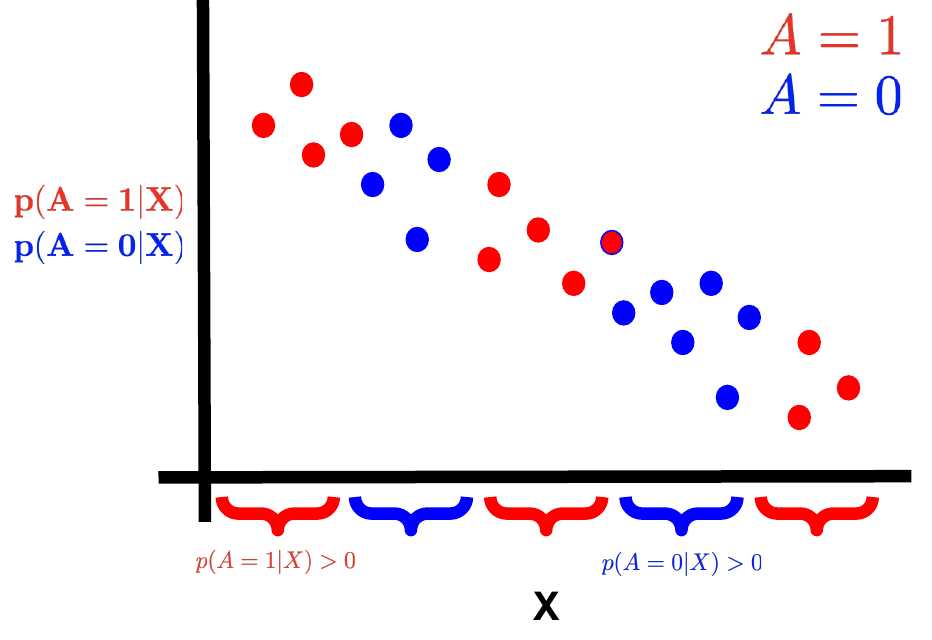}
        \subcaption{Generalization through interpolation}
        \label{fig:interpolation}
    \end{minipage}
    \caption{Generalization of the propensity score in two scenarios where the positivity assumption is violated. (a) requires extrapolation and
    (b) requires interpolation for generalizing to unseen counterfactual examples respectively.}
    \label{fig:genearlization_positivity}
\end{figure}

\paragraph{Positivity via generalization.}

Positivity is hard to satisfy in a strict sense, as we must have as many data points as there are actions for each and every possible $x$ with $p(x)$. This is largely impossible when $x$ is continuous, and even when $x$ is discrete, it is difficult if the support of $p(x)$, i.e., $\left\{ x \in \mathcal{X} | p(x) > 0 \right\}$ is large. 
Instead, we can fit a parametric or non-parametric model to 
model $p(A=a|X)$ that can generalize to an unseen combination of $(X,A)$.
In such a case, 
the success of generalization depends on the support of data points we have access to.
Figure~\ref{fig:genearlization_positivity} presents a depiction of two cases where the fitted model must respectively extrapolate and interpolate. 
In the latter case of interpolation, even without positivity, we may be successful at correctly inferring the causal effect, but in the former case, it would be much more challenging. This suggests we must be careful at relying on generalization to overcome the issue of positivity (or lack thereof.)

\newpage
\section{Active Causal Inference Learning}

\begin{algorithm}[t]
\caption{Active CI protocol}\label{alg:active_ci}
\begin{algorithmic}
    \State $A$ actions, $T$ rounds (both known); potential outcome $Y(a)$ for each action $a$ (unknown {\it a priori}).
    \While {each round $t \in [T]$}
        \State Observe a covariate $x_t$
        \State Pick an action according to policy, $a_t\sim\pi(x)$.
        \State Outcome observed $y_t \in [0,1]$ is sampled given $X=x_t,A=a_t$.
        {\color{blue}
        \If {action $a$ sampled from $\pi(x) = p(A|X=x)$}
            \State Update the potential outcome $\mathbb{E}_{p(X)}[Y_X(a)] \leftarrow t^{-1}\sum_{t'=1}^t \frac{\mathbb{I}(a_{t'})}{p(a|x)} y_t$
        \Else
            \State Update the potential outcome $\mathbb{E}_{p(X)}[Y_X(a)] \leftarrow t^{-1}\sum_{t'=1}^t \frac{\mathbb{I}(a_{t'})}{p(a)} y_t$
        \EndIf
        }
        \State [Optional] Update the policy $\pi$.
    \EndWhile
\end{algorithmic}
\end{algorithm}

Because counterfactual information is rarely available together with factual information in observed data \citep{Graepel2010, Li2010, Chapelle2015}, one approach to causal inference (CI) is to design an online algorithm that actively explores and acquires data. 
We introduce and describe an active CI framework that combines causal inference and data collection, inspired by works of literature on contextual bandit\citep{Slivkins2019, Bouneffouf2020}.
In this framework, an algorithm
estimates the expected potential outcome $Y_X(A)$ of each action and collects further data for underexplored counterfactual actions. A general active CI protocol is presented in Algorithm~\ref{alg:active_ci}. 
We denote $(x_t,a_t,y_t)$ for observed covariate, action, and outcome at time $t$, and denote $Y_X(A)$ for potential outcome. We update the estimated expected potential outcome $\mathbb{E}_X[Y_X(A)]$ based on a new observed data point $(x_t,a_t,y_t)$.
Although active CI algorithms are inspired by contextual bandit, there is a major difference between these two; that is, CI focuses much more on 
estimating the expected potential outcomes,
while bandit algorithms on finding an optimal policy function that maximizes the expected outcome values.\footnote{We use {\em outcome} and {\em reward} interchangeably.}

Most of the time CI researchers and practitioners face the challenge of being limited to data exploration not by our choice, but by other factors that make it inaccessible to A/B testing from real-world constraints. 
Active CI learning benefits from the policy function $\pi(x)$ when it can compromise between exploration and exploitation for understanding the causal effect and optimizing the decision in real-world applications respectively.
In this section, we review active CI literature by first examining RCT and then expanding it to contextual bandit methods.

\subsection{Randomized Controlled Trial}
\label{sec:rct}

Randomized controlled trial (RCT) is the most well-known and widely practiced CI method. It is {\it de facto} standard in e.g. clinical trials, where participants are  randomly divided into treatment and control groups, and the outcomes of these two groups are used to compute the average treatment effect (ATE).
RCT falls under the active CI method group since it collects data proactively by randomly assigning an action to each data point \citep{rubin1974estimating}. 
Although it is typical to choose a uniform distribution over an action set $\pi = \mathcal{U}[\mathcal{A}]$ as a policy function in RCT, we can in principle pick any random distribution as long as it is independent of the covariate $\pi=\mathcal{P}[\mathcal{A}]$, as shown in Algorithm~\ref{alg:rct}.\footnote{
    Existing literature often conflates having an equal chance of sampling each action with independence, but this is not true.
}

\begin{algorithm}[h!]
\caption{Randomized Controlled Trial}\label{alg:rct}
\begin{algorithmic}
        \State Observes a covariate $x_t$.
        \State Pick an action according to a policy, {\color{blue}$a_t\sim\pi=\mathcal{P}[\mathcal{A}]$}.
        \State Observe the outcome $y_t \in [0,1] \sim Y_{X}(A) | X=x_t,A=a_t$.
        \State Estimate $\mathbb{E}_{p(X)}[\hat{Y}_{X}(a)]$ for all $a$.
\end{algorithmic}
\end{algorithm}

A set of points $\lbrace (x_t, a_t, y_t)\rbrace^T_{t=0}$ collected by RCT automatically satisfy the exchangeability assumption.
Because actions are selected independently of covariate $X$ as shown in Figure~\ref{fig:condition_intervention_graph},
the potential outcome $Y_X[A]$, estimated from this data, must be preserved even if action $P(A|X)$ changes.
With these we can show that the conditional and intervention probabilities coincide with each other, allowing us to perform causal inference from data collected by RCT:
\begin{align*}
    p(Y=y|do(A=a)) &= \sum_x p(Y=y|A=a,X=x)p(x)\\
        &= \sum_x \frac{p(Y=y|A=a,X=x)p(A=a|X=x)p(x)}{p(A=a|X=x)} \\
        &= \sum_x \frac{p(Y=y,A=a,X=x)}{p(A=a|X=x)}\\
        &= \sum_x \frac{p(Y=y,A=a,X=x)}{p(A=a)} \text{ (since $A \indep X$) }\\
        &= \sum_x  p(Y=y,X=x|A=a)\\
        &= p(Y=y|A=a). \numberthis \label{eqn:rct_proof}
\end{align*}    

Let us consider Simpson's paradox (Ex.~\ref{ex:simpson_paradox} in \S\ref{sec:assumption_discussion}).
Although there was an overall strong association between being a man and being more likely to be admitted, 
we found different associations when we considered different sub-populations 
due to the uneven sex distribution among the applicants and the acceptance rates, across departments. 
With RCT using the uniform action distribution, we would end up with an even number of each sex independent of the department choice, i.e., $p(A=a|S=\text{man})$ = $p(A=a|S=\text{woman})$.
This would allow us to verify whether men are more likely to get admitted to graduate school without being confounded by the department's choice. 

It is often tedious and difficult to plan and design an RCT due to many biases that are difficult to mitigate. One such example is a control bias which arises when 
the control group behaves differently or is treated differently from the treatment group.
For instance, participants in the control group may be more likely to drop out of the study than those in the treatment group, due to the lack of progress they perceive themselves. 
In order to avoid a control bias, one must carefully consider eligibility criteria, selection of study groups, baseline differences in the available population, variability in indications in covariates, and also the management of intermediate outcomes \citep{Simon2001IsTR, Janewit2010}.
There are techniques that help you assess the quality of RCT studies with an emphasis on measuring control bias in the planning and implementation of experiment designs \citep{Chalmers1981, Chalmers1989SelectionAE, Stewart1996BiasIT, Moreira2002, Olivo2008ScalesTA}.

RCT is widely used in clinical trials in medicine \citep{Concato2000, Rothwell2005ExternalVO, Green2006EvaluatingTR, Frieden2017EvidenceFH}, policy research in public health \citep{Sibbald1998UnderstandingCT, MartelGarca2010TheoryEV, Deaton2016UnderstandingAM, Choudhry2017RandomizedCT}, econometrics \citep{HECKMAN1985239, LaLonde1986, Banerjee2012} and advertisements in marketing \citep{Graepel2010, Chapelle2015, Gordon2019}.
In these real-life applications, we often cannot afford to blindly assign actions but to determine the action based on the covariate $X$, due to ethical, legal and economical reasons.  
Because it is challenging to apply RCT in practice \citep{Saturni2014RandomizedCT}, some studies explore combining RCT with observational study data \citep{rubin1974estimating, Concato2000, Hannan2008}.  For example, one can use inverse probability weighting or matching techniques to de-bias the potential outcome estimation, as we will discuss in \S\ref{sec:passive_cil}. 
We thus review active CI methods with a covariate-dependent data collection policy in the rest of this section.

\subsection{Causal inference with contextual bandits}

It is often impractical to use RCT in real-life settings due to (but not limited to)
the following three limitations \citep{rubin1974estimating, Olivo2008ScalesTA, Schafer2009, Saturni2014RandomizedCT}.
First, a sample size must be large enough to detect a meaningful difference between the outcomes of actions, due to the high variance of RCT.
Second, complete randomization or complete independence from the covariate is often neither feasible nor ethical in practice.
Lastly, complete randomization often goes against 
the real-world objective of maximizing the outcome which is different from correctly inferring the causal effect.
For example, doctors should not randomly assign different treatments to patients in order to test their causal effects on the patients, because this could end up harming many patients. 
Instead, a doctor makes the best decision for each patient given their expertise (i.e. their own policy),
and on the fly
adjusts their policy online in order to maximize the outcome of each patient:
\begin{align*}
    \arg\max_{\pi_e} \mathbb{E}_{x\sim p(X)} \mathbb{E}_{a\sim \pi_{e(x)}}\left[Y_X(a)\right].
\end{align*}
RCT on the other hand does not maximize the outcome at all, which makes it less desirable to use in many real-world scenarios.


In this section, we review different ways of performing both tasks together, that is, finding an optimal policy and estimating causal effects.
In particular, we examine various ways to intervene and actively collect data under the framework of contextual bandits.\footnote{
See Appendix~\ref{app:bandits_algo} for the basic description of contextual bandits.
}
There are two primary approaches to applying contextual bandit algorithms to CI. 
The first approach actively gathers interventional data 
and then estimates ATE from this actively collected randomized dataset.\footnote{
A randomized dataset refers to a dataset consisting of tuples collected using actions chosen independently of covariates.
} 
The second approach uses a contextual bandit method to learn a causal model utilizing all the collected data points including both randomized and non-randomized actions.


\subsubsection{Estimating ATE from interventional data collected with a bandit method}
\label{app:bandits_ci1}

The general idea is to keep track of data points for which intervention, i.e. randomization, happened. 
It is no different from RCT except that random intervention happens only occasionally based on your choice of bandit algorithm.
For the purpose of illustration, we use the $\epsilon$-greedy strategy together with the expert's policy as an example in Algorithm~\ref{alg:active_ci_eps_experts}. 
The epsilon greedy algorithm is an easy way to add exploration to the basic greedy algorithm; 
we greedily choose an action based on the estimated highest outcome values but 
once in a while randomly select an action independently of the covariate with the probability $\epsilon$. 
We use a {\it randomized set} to refer to a collection of these randomized actions together with associated outcomes and covariates.
We then estimate the average treatment effect (ATE) directly using the randomized dataset only (see Appendix~\ref{chp1:agg}).
The efficiency in estimating the causal effect is determined solely by how often we randomize action, i.e., the probability $\epsilon$. 
In contrast to RCT, as we select actions based on the covariate-aware policy occasionally with the probability $1-\epsilon$, 
we fulfill both outcome maximization and ATE estimation, although the efficiency in ATE estimation is typically worse than that of RCT.

\begin{algorithm}[t]
\caption{$\epsilon$-greedy protocol}\label{alg:active_ci_eps_experts}
\begin{algorithmic}
    \State $A$ actions, $T$ rounds (both known); potential outcome $Y(a)$ for each action $a$ (unknown).
    \While {In each round $t \in [T]$}
        \State {\color{blue}Toss a coin with the exploration probability $\epsilon_t$.}
        \If {{\color{blue}explore}}
            \State explore: choose an action $a_t \sim \mathcal{U}[a]$
        \Else
            \State Observe a covariate $x_t$
            \State Pick an action according to {\color{blue} the expert $a_t\sim\pi_e(x)$}.
        \EndIf
        \State Observed the outcome $y_t \sim  Y | X=x_t, A=a_t \in [0,1]$.
        \State {\color{blue} Store $y_t$ in set $\mathcal{D}$ if explore}
        \State Update the expected potential outcome ${\color{blue}\mathbb{E}_{p(X)}[Y_X(a)] \leftarrow |\mathcal{D}|^{-1}\sum^{\mathcal{D}}_d \mathbb{I}[A=a] y_d }$ for all $a$.
    \EndWhile
\end{algorithmic}
\end{algorithm}

There are other, more sophisticated methods such as high-confidence elimination and upper-confidence bound (UCB) algorithms \cite{Auer2002b, Slivkins2019}.\footnote{
See Appendix~\ref{app:bandits_algo} for more details.
}
Unlike the $\epsilon$-greedy strategy, these approaches choose when to randomize action based on a learned policy so far. 
A {\it high-confidence elimination method} alternates between two actions, $a$ and $a^\prime$ until the confidence bounds of the two actions' potential outcomes do not overlap, where the confidence bounds are defined as 
\begin{align*}
    UBC_t(a) = \mathbb{E}\left[Y(a)\right] + r_t(a)\\
    LBC_t(a) = \mathbb{E}\left[Y(a)\right] - r_t(a)
\end{align*} 
with 
the confidence radius $r_t(a)=\sqrt{2\log(T)/n_t(a)}$\footnote{
There are ways to estimate a tighter radius based on different assumptions according to the previous research \cite{Slivkins2019}.
} 
and the number $n_t(a)$ of rounds with the action $a$\cite{Slivkins2019}.

Once the lower-confidence bound of the potential outcome of one action is greater than the upper-confidence bound of that of the other action, i.e., $UCB(a) < LCB(a^\prime)$, the former action $a^\prime$ is selected indefinitely from there on because the abandoned action cannot be the best action in terms of maximizing the outcome value.
The rationale behind this method is to make sure to explore until we are confident about the expected potential outcome of each action.
In short, the high-confidence elimination method fully explores until the best action is determined with a high level of confidence, after which it converts to exploiting the discovered best action only. 
The first phase of exploration is thus similar to performing RCT.
This approach is interesting because as soon as the lower bound of $a'$ and the upper bound of $a$ separate, we automatically get some level of assurance about the potential outcome estimates.

The {\it UCB algorithm} on the other hand picks an action that maximizes $UCB_t(a)$ at every round $t$.
This choice automatically balances exploration and exploitation. 
$a^\prime$ is selected over another action $a$ if $UCB_t(a^\prime) > UCB_t(a)$, which can happen for one of two reasons; the uncertainty is high (exploration) and the actual reward is high (exploitation). 
For the purpose of estimating ATE, we only use the randomized (explored) data points where the confidence radius $r_t(a)$ was large.
There are other methods such as Thompson-sampling causal forest \cite{Dimakopoulou2017} and random-forest bandit \cite{Feraud2016}, which share a similar flavour with the UCB algorithm.

\subsubsection{Learning a causal graph using a bandit algorithm}
\label{app:bandits_ci2}

So far in this section, we have discussed how to estimate ATE directly by gathering an interventional dataset using bandit methods. 
While such an approach allows us to measure the expected potential outcomes and ATE, we cannot infer conditional potential outcomes.
One of the CI goals is to measure the causal effect using ATE but a bigger and more ambitious goal is to answer counterfactual questions. The latter is only possible if we can infer individual potential outcomes.
Here, we discuss how to learn an underlying graphical model using a contextual bandit method in order to infer individual potential outcomes.

\begin{figure}[t]
    \begin{minipage}{0.28\linewidth}
        \centering
        \includegraphics[width=\linewidth]{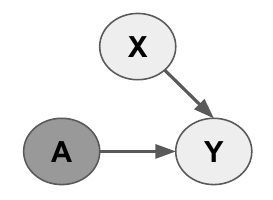}
        \subcaption{Unconfounded graph}
        \label{fig:rct}
    \end{minipage}
    \begin{minipage}{0.23\linewidth}
        \centering
        \includegraphics[width=\linewidth]{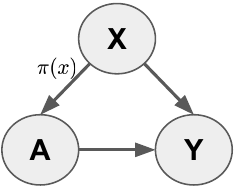}
        \subcaption{Contextual Bandit}
        \label{fig:contextualbandit}
    \end{minipage}
    \begin{minipage}{0.23\linewidth}
        \centering
        \includegraphics[width=\linewidth]{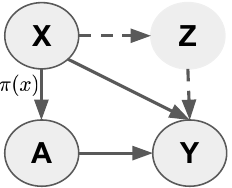}
        \subcaption{SemiparametricBandit}
        \label{fig:semiparametric_bandits}
    \end{minipage}
    \begin{minipage}{0.23\linewidth}
        \centering
        \includegraphics[width=\linewidth]{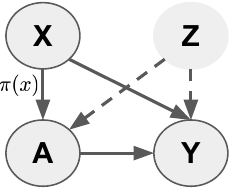}
        \subcaption{Causal Bandit}
        \label{fig:casual_bandit}
    \end{minipage}
    \caption{Bandit methods for Active CI}
    \label{fig:active_ci}
\end{figure}

At each trial $t$, we approximate the potential outcome $\hat{Y}_{X}(a_t)$ using a parametrized model $g_\theta(x_t,a_t)$ that computes the outcome based on the context $x_t$ and action $a_t$,
\begin{align}
    \hat{Y}_{X}(a_t) = g_{\theta}(x_t, a_t) + \epsilon_t,
    \label{eqn:value_function}
\end{align}
where $\epsilon_t \sim \mathcal{N}(0,\sigma^2)$ is noise which comes from unknown confounders at time $t$.
Although data was collected by 
the generic probabilistic graph that includes a policy $\pi$ in Figure~\ref{fig:contextualbandit},
$g_\theta$ learns a causal graph in Figure~\ref{fig:rct}, because
the parameters $\theta$ are estimated from the intervention dataset alone.
This allows us to use $g_\theta$ to infer the potential outcome of a counterfactual action and/or of unseen covariates.

Contextual bandits can handle causal graphs that are more complicated. For instance, in Fig.~\ref{fig:semiparametric_bandits}, there is an extra variable $Z=f(X)$ that mediates the effect of the covariate $X$ on the outcome $Y$ but does not affect the action $A$.  
In this case, the outcome function is in the form of 
\begin{align*}
    \hat{Y}_X(a_t) = g_\theta(x_t,a_t) + f_\phi(x_t) + \epsilon_t,
\end{align*}
where $g_\theta(x_t,a_t)$ takes into account the confounder $X$ as well as policy function $\pi(X)$ and maps them to an output $Y$. 
$f_\phi(X)$ learns to isolate the effect of the covariates that affect the outcome independent of the action.
$\epsilon_t$ is the noise.
Having $f_\phi(X)$ to learn covariates that are not confounders can reduce the complexity of learning for $g_\theta(x_t,a_t)$. 

When both $g_\theta$ and $f_\phi$ are non-parametric, it is often computationally intractable to tackle this problem.
In order to avoid these issues of computational tractability and undesirable regret bound, {\em semiparametric contextual bandits} consider parametric policies \citep{Krishnamurthy2018, Greenewald2017, Peng2019} and learn a linear bandit model using regression oracles to estimate potential outcome \citep{Swaminathan2017} 
\begin{align*}
    \hat{Y}_X(a_t) = w^\top x_t + f_\phi(x_t) + \epsilon_t,
\end{align*}
where $w^\top$ is a parameter vector and $\epsilon_t$ is noise.
They have a nice property where the regret bound is the same as the regular contextual bandit's regret. 
The action-independent features $f_\phi(x_t)$ get cancelled out in the regret formulation together with the noise term $\epsilon_t$, because they are independent of the action choice.

\subsubsection{Correcting a bias from unobserved confounders}

It is unrealistic to observe all confounders in practice, nor for us to verify whether all confounders have been observed.  
Any confounding factor not included in the covariate leads to difficulties in accurately estimating the causal effect. 
Here, we consider a way to detect and correct such a bias that arises from having unobserved confounders (see Figure~\ref{fig:casual_bandit}).

Consider a policy function $\pi^*(x')$ that is optimized for maximizing $\arg\max_{a} CATE(x',a)$:\footnote{
Unlike $CATE(x)$ with a binary action in Equation~\ref{eqn:cate_bianry}, we define $CATE(x,a)$ for a set of more than two actions.
}
\begin{align*}
    \pi^*(x') = \arg\max_{a} CATE(x',a) = \arg\max_{a} \mathbb{E}_{Y, X\backslash X'}\left[Y_X(a)- \max_{\bar{a}}{Y_X(\bar{a})}|X'=x'\right].
\end{align*}  
This $\pi^*(x')$ is estimated without considering unobserved confounders like in Figure~\ref{fig:contextualbandit}, and yet, with the realization of unobserved confounders shown in Figure~\ref{fig:casual_bandit}, we now correct the estimation.
Let $\bar{a}$ be the counteraction to $\pi^*(x')$ that returns a larger CATE than the originally selected action $\pi^*(x')$.
Suppose we found $x'$ where $CATE(x',\pi^*(x')) < CATE(x',\bar{a})$ where $\bar{a} \neq \pi^*(x')$. 
Since the policy function $\pi^*$ is optimal, this can only happen if there are unobserved confounders.
This illustrates that our learned policy might not be the best in such a situation.
We then need to search for a new policy function $\pi'$ that can handle this situation better.

Because our counteraction could not be better than the action $\pi^*(x')$ without unobserved confounders, we compare the potential outcomes of the action and the counteraction $\bar{a}$;
\begin{align}
    \mathbb{E}_{Y,X\backslash X'}\big[Y_X(\pi^*(x'))| X'=x' \big] \text{ and }  \\
    \mathbb{E}_{Y,X\backslash X'}\big[Y_X(\bar{a}) | A=\pi^*(x')|X'=x'\big], 
    \label{eqn:bandit_unobserved_confounders}
\end{align}
respectively.
In order to compute the outcome of the counteraction, we actively intervene with the counteraction and collect a new sample 
$(x', y^\prime, \bar{a}, a)$.
We also count the frequency of the expected potential outcome of counteraction being higher than action $\pi^*(x')$,
\begin{align*}
    f(\pi^*) = \sum^N_i \mathbb{I}\left[\mathbb{E}_{Y,X\backslash X'}\left[Y_X(\pi^*(x^\prime_i))| X'=x^\prime_i\right] > \mathbb{E}_{Y,X\backslash X'}\left[Y_X(\bar{a})| A=\pi^*(x^\prime_i), X'=x^\prime_i\right]\right],
\end{align*}
where $\frac{f(\pi^*)}{N}$ tells us how often $\pi^*$ is wrong.
Our new policy function $\pi^\prime(x')$ samples a new action according to the Bernoulli distribution with probability $\frac{f(\pi^*)}{N}$. This strategy copes with the fact that our policy is biased and incorrect with the probability $\frac{f(\pi^*)}{N}$ due to unobserved confounders. 

\citet{Bareinboim2015} show that while this method requires collecting enough data for $f$ to converge, we can optimize a new policy function faster by weighting the samples from the Beta distribution $\mathcal{B}(f,(1-f))$ by the bias as shown in Algorithm~\ref{alg:causal_ts}. 
The bias from the unobserved confounder is defined as one minus the absolute treatment effect \citep{Bareinboim2015}, 
\begin{align}
    \text{bias} = 1 - |\mathbb{E}_{Y,X\backslash X'}[Y_X(\bar{a})| A=\pi^*(x'), X'=x'] - \mathbb{E}_{Y,X\backslash X'}[Y_X(\pi^*(x'))| X'=x'] |.
    \label{eqn:causal_bandit_bias}
\end{align} 
Eq.~\ref{eqn:causal_bandit_bias} quantifies how effective applying action $\pi^*(x')$ over the counteraction $\bar{a}$ is. 
If this quantity is large, the re-weight of the probability of choosing action $\pi^*(x')$ by the bias remains high. 
Otherwise, the re-weight of the probability of choosing counteraction $\bar{a}$ by bias remains high. 
Overall this encourages faster convergence of re-estimating the potential outcome (see the last three lines of Algorithm~\ref{alg:causal_ts}). 
This specific algorithm is called causal Thompson sampling.

\begin{algorithm}[t]
\caption{Causal Thompson Sampling (TS$^\text{C}$) \citep{Bareinboim2015} }\label{alg:causal_ts}
\begin{algorithmic}
    \State Let $a=\pi^*(x')$ be intuitive action.
    \State Let $\bar{a}$ be counteraction to $a$.
    \State Let $\mathbb{E}_{Y,X\backslash X'}[Y_X{A=a}|A=a,X'=x']$ be the expected payout for intuitive action
    \State Let $\mathbb{E}_{Y,X\backslash X'}[Y_X{A=\bar{a}}|A=a,X'=x']$ be the expected payout for counter-intuitive action\\
    
    \While {$t=1 \cdots T$}
        \State Let $w = [1,1]$
        \State Sample $a \sim \text{intuition}(x_t,t)$ \\

        \State // Estimate the potential outcomes and bias
        \State Compute $bias = 1 - | \mathbb{E}[Y_X(\bar{a})|A=a,X'=x'] - \mathbb{E}[Y_X(a)|X'=x'] | $ (Equation~\ref{eqn:causal_bandit_bias})
        \If{$\mathbb{E}[Y_X(\bar{a})|A=a,X'=x']  > \mathbb{E}[Y_X(a)|X'=x']$}
            \State Set $w[a] = bias$
        \Else 
            \State Set $w[\bar{a}] = bias$
        \EndIf\\

        \State // Choose a new policy $\pi'(x')$ based on new weighting
        \State $\beta_1 = \mathcal{B}(f(a), (1-f(a)))$
        \State $\beta_2 = \mathcal{B}(f(\bar{a}), (1-f)(\bar{a}))$
        \State Set $\dot{\pi}(x') \leftarrow \max(\beta_1 \cdot w[a], \beta_2 \cdot w[\bar{a}])$
        \State Set $\dot{y} = \text{simulate}(\dot{\pi}(x'))$
        \State Update $\mathbb{E}[Y_X(\dot{\pi}(x'))|A=a,X'=x'] \text{ with } (x', \dot{y}, \dot{\pi}(x'), a)$
    \EndWhile
\end{algorithmic}
\end{algorithm}

\newpage
\section{Passive Causal Inference}
\label{sec:passive_cil}

Unlike in active causal inference (CI), in passive CI, we must calculate ATE given a non-randomized dataset which was gathered in advance with the actions chosen based on covariates \citep{rubin1974estimating, Holland1986}. 
As discussed earlier in \S\ref{sec:assumption_discussion}, we assume that the dataset satisfies positivity, although we discuss how to relax it with deep learning later in this section. 
In this section, we present and examine passive approaches to CI, which are gaining more popularity. Most of them are primarily grounded in one of the following three general approaches: matching, inverse probability weighting and doubly-robust methods. We first review these basic approaches and introduce some of the more recent approaches.

\subsection{A Naive Estimator}
\label{sec:naive_estimator}

Before we begin, we start with a naive estimator $\mu_A(X)$ of the outcome variable $Y$.
Let $\mu_A: \mathcal{X} \rightarrow \mathcal{Y}$ be the generic estimator that predicts the outcome.
For example, this estimator can be simple empirical averaging: 
\begin{align*}
    \mu_a(x) = \frac{\sum^{N}_{i=1} y_i \mathbb{I}[a_i=a,x_i=x]}{\sum^{N}_{i'=1} \mathbb{I}[x_{i'}=x]}, 
\end{align*}
where $D=\lbrace (x_i, a_i, y_i) \rbrace^{N}_{i=1}$ is a dataset that consists of $N$ data points, and $\mathbb{I}$ is an indicator function. This estimator looks at the average outcome of the action $a$ given a particular covariate $x$.
Another example would be to have a parametrized estimator, $\mu_A(X; \theta)$ where $\theta$ is a parameter. 

Such a naive estimator $\mu_A(X)$, which often maximizes the log-likelihood, is a biased estimator of the potential outcome $Y$, due to the discrepancy between the conditional and invention probabilities.
In the subsequent sections, we introduce and discuss modifications either to the estimator or to the dataset, that allows us to obtain an unbiased estimate of the potential outcome.

\subsection{Matching}
\label{sec:matching_method}

ATE estimation is hard when working with a real-life dataset due to 
confounding.
A common approach is to construct a randomized dataset, that is free of confounding, from a non-randomized one. 
The matching method achieves this by pairing each treated instance with a control instance and ignoring any unmatched instance. This process
balances the treatment and controlled group \citep{Scotina2019}.

Ideally, we should have both factual and counterfactual outcomes for every data point ($\langle (x_i, y_{i}(1),y_{i}(0) \rangle$ for all $i$).
This is often impossible due to the fundamental problem of CI, where the problem arises from the very fact that we can only observe the outcome of a single action given a particular covariate.
We instead aim for approximate one-to-one matching where we pair treated data with controlled data that are similar enough.
That is, we pair two instances if 
$\lbrace \langle (x_i, 1, y_{i}(1)), (x_j,0, y_{j}(0)) \rangle | D(x_i,x_j) < \epsilon \text{ for } i\neq j \rbrace $ and for some small $\epsilon$, where $D(\cdot,\cdot)$ is a problem-specific distance metric. 
We then compute ATE using our new balanced dataset.
Matching methods remove confoundedness by creating comparable groups based on observed covariates.  It selects individuals from different treatment groups but has similar characteristics or covariate distributions. This ensures that the treatment and control groups are balanced concerning the potential confounders.
The advantages and disadvantages of such a matching method lie in the bias-variance tradeoff.
The advantage of the matching method is that it reduces the confounding bias, but it increases the variance because we removed (potentially many) unmatched data points.

There are many standard metrics of {\em closeness} that are widely used. 
One such metric is Mahalanobis distance:
\begin{align*}
    D_{ij} = (x_i-x_j)^\top \Sigma^{-1}(x_i-x_j), 
\end{align*}
where 
$\Sigma$ is the covariance metric.
Many alternative metrics have been proposed over the past decades, such as those relying on a coarsened data space or other feature spaces~\citep{Cochran1973, Rubin1979UsingMM, ROSENBAUM1983, RUBIN1992, Rubin2006, Stuart2010, Stefano2012, Zubizarreta2012, Zhao2004, Resa2016}. The choice of a metric must be determined for each problem separately.

Matching methods have evolved from a greedy algorithm based (heuristic-based search) to optimal/full matching \citep{Kim2016}.
The greedy method ends up with a sub-optimal solution since the ordering of pairing matters \citep{Weitzen2004}.
Moving away from greedy matching, 
one can use an optimal non-bipartite matching algorithm that runs in polynomial time \citep{Bo2004, Dehejia1999}. 
Such an algorithm generates a series of matched sets that consist of at least one treated individual and at least one control individual. 
It is optimal in terms of minimizing the average distance between each treated individual and each controlled individual within each matched set \citep{Hansen2004}.
There are other approaches such as weighting methods where one adjusts the importance of distance between data points \citep{Heckman1997, Hirano2003, imbens2004nonparametric}. These methods can be helpful when the data samples are unevenly spread out over the domain.

There however remain challenges with matching. 
First, it is difficult to have exact matches for all data points with a non-randomized or imbalanced dataset (the dataset is unevenly distributed w.r.t actions).
We thus end up eliminating a significant number of unpaired data points after matching between treated and controlled groups, resulting in the loss of information and statistical power. 
There are however some algorithms that enable many-to-one matching, with the simplest being the $K$-nearest neighbour method
\citep{Karp1972, Schafer2009, Zubizarreta2012}. 
Second, matching works well when the dimensionality of the covariate is low, but it easily fails when the dimensionality is high due to the curse of dimensionality \citep{Gu1993ComparisonOM}. It can be helpful to use deep learning to obtain a more concise and dense representation, similar to what we will discuss in \S\ref{sec:ci_representation_learning}.

\subsection{Inverse Probability Weighting}
\label{sec:ipw}

Inverse probability weighting (IPW) removes the effect of confounders by weighting the potential outcome of each action by its inverse probability weight \citep{ROSENBAUM1983, Robins1994, Hirano2003}:
        \begin{align}
            \mathbb{E}\left[\frac{Y\mathbb{I}[A=a]}{p(A|X)}\right] &= \mathbb{E}_{p(X)}\left[ \mathbb{E} \left[ \frac{Y(a)\mathbb{I}[A=a]}{p(A|X)} \Big| X\right] \right] 
                = \mathbb{E}\left[ \frac{ \mathbb{E} \left[ Y(a) \Big| X\right]\mathbb{E} \left[ \mathbb{I}[A=a|X] \right]}  {p(A|X)} \right]
                = \mathbb{E}\left[ Y(a) \right].
                \label{eqn:ipw_estimator}
        \end{align}
        Equation~\ref{eqn:ipw_estimator} illustrates that we can take the subset of data that corresponds to a particular action $a$ as long as we can divide by the so-called propensity score in order to compute $\mathbb{E}[Y(A)]$.
        $p(A=a|X=x)$ is known as the {\em propensity score} $e(x)$.\footnote{
        Additionally, propensity scores can be used for matching methods where one compares two covariates with similar propensity values $D(\hat{e}(x_i), \hat{e}(e_j))$ for $i\neq j$ and some distance metric 
        \begin{align*}
            D_{ij} = |e(x_i) -e(x_j)|,
        \end{align*}
        where $e(x_i)$ and $e(x_j)$ are the propensity scores for the data point $x_i$ and $x_j$, respectively.
        The propensity score is a popular method as it summarizes the entire covariate into a single scalar
        and has been shown to be effective in theory \citep{ROSENBAUM1983, RUBIN1992, Rubin2006, Zubizarreta2012, Diamond2013, Resa2016, Abadie2016} and in practice \citep{Jalan2001, Dehejia2002, Monahan2011, Amusa2018}.
        } 

        The propensity score theorem \citep{ROSENBAUM1983} tells us that if ignorability is satisfied given $X$, then ignoreability conditioned on $e(X)$ is also satisfied \citep{Imbens2015}:
        \begin{align*}
            (Y(1),Y(0)) \indep A | X \implies (Y(1),Y(0)) \indep A | e(X).
        \end{align*}
        This predicate tells us that the potential outcomes are independent of $A$ given confounder $X$,
        and this is due to the blocking back-door criterion conditioning on confounder $X$ \citep{Pearl09a}.
        Similarly, the potential outcomes are independent of $A$ given the propensity score $e(x)$, and the propensity score has the same effect as removing the blocking back-door path in a causal graph  by conditioning on the edge between $X$ and $A$. 
This illustrates that the 1-dimensional score function, that is the propensity score, is enough to compress the high-dimensional confounder $X$.

        While simple finite-sample averaging $\widehat{ATE}_{AGG}$ is a biased estimator of ATE,
        \begin{align*}
            \widehat{ATE}_{AGG} &=  \frac{1}{N}\sum_{i}^{n}\left[\frac{\mathbb{I}[A_i=1]y_i(1)}{\eta(x_i)}\right] - \frac{1}{N}\sum_{i}^{n} \left[\frac{\mathbb{I}[A_i=0]y_i(0)}{1-\eta(x_i)}\right],
        \end{align*}
        $\widehat{ATE}_{IPW}$ is an unbiased estimator of ATE,
        \begin{align*}
            \widehat{ATE}_{IPW} &=  \frac{1}{N}\sum_{i}^{n}\left[\frac{\mathbb{I}[A_i=1]y_i(1)}{e(x_i)}\right] - \frac{1}{N}\sum_{i}^{n} \left[\frac{\mathbb{I}[A_i=0]y_i(0)}{1-e(x_i)}\right],
        \end{align*}
        where  $eta(x)=\frac{N_{x1}}{N}$ and $N_{x1}$ is the number of treated samples.
        Note that $\widehat{ATE}_{AGG}$ uses a naive estimator from \S\ref{sec:naive_estimator} to estimate ATE.
        $\widehat{ATE}_{IPW}$ is known as {\em an oracle IPW estimator} since the propensity score $e(\cdot)$ is known.
        The estimation error, defined as $\sqrt{N} |ATE - \widehat{ATE}_{IPW}|$, follows 
        a Normal distribution with zero mean and the variance of
        \begin{align}
            \text{VAR}_\text{IPW} = \text{Var}[ATE_\text{AGG}(X)] 
            + \mathbb{E}_p(x)\left[ \frac{\sigma^2(X)}{e(X)(1-e(X))} + \frac{c(X)^2}{e(X)(1-e(X))}\right],
            \label{eqn:variance_IPW}
        \end{align}
        where 
        $c(X)$ is a function that satisfies 
        \begin{align*}
            Y(0) &= c(X) - (1-e(X))ATE(X) + \epsilon(X)\\
            Y(1) &= c(X) + e(X)ATE(X) + \epsilon(X),
        \end{align*}
        where $\epsilon(X)$ is a Gaussian noise with variance of $\sigma^2(x)$.
        $\widehat{ATE}_{AGG}$ can be seen as a version of the IPW estimator with an imperfect propensity score being $\hat{e}(x)=\frac{N_{x1}}{N}$. 
        $\text{VAR}_\text{IPW}$ demonstrates that even with using the true propensity score $e(x)$, $\widehat{ATE}_{IPW}$ has a worse asymptotic variance than $\widehat{ATE}_{AGG}$ which can be thought of as inverse probability reweighting with an incorrect propensity score $\hat{e}(x)$.
        This is an example of the bias-variance trade-off. 
        

        A plethora of methods have been proposed since then, that are unbiased and exhibit lower variance than the oracle IPW above by replacing the propensity score with other weighting schemes $\hat{e}'$.
        For example, one can reduce the variance of an IPW estimator by normalizing the weights \citep{Hirano2003}:
        \begin{align*}
            \widehat{ATE}_{SW} =  \frac{\sum_{i}^{n}\mathbb{I}[A_i=1]y_i(1)w(x_i)}{\sum_{i}^{n}\mathbb{I}[A_i=1]w_1(x_i)} - \frac{\sum_{i}^{n} \mathbb{I}[A_i=0]y_i(0)w_0(x_i)}{\sum_{i}^{n} \mathbb{I}[A_i=0]w_0(x_i)},
        \end{align*}
        where $w_1 = \frac{1}{e(x_i)}$ and $w_0=\frac{1}{1-e(x_i)}$. This leads to a lower variance in the estimate, and these weights are thus called {\em stabilized weights} \citep{Robins2000}.
        According to \citet{Hirano2003}, $\widehat{ATE}_{SW}$ outperforms $\widehat{ATE}_{IPW}$ in terms of asymptotic convergence rate \citep{Hirano2003}. 
        \citet{Lunceford2004} review various versions of the IPW estimator and suggest a way to take into account the uncertainty in estimating the propensity score using a closed-form sandwich estimator (M-estimator \citep{Stefanski2002}) \citep{Lunceford2004}.

\subsection{Doubly Robust Methods}
    \label{sec:doubly_robust}

    It is often hard to obtain the propensity score in advance nor guarantee that our propensity estimate $\hat{e}(x)$ is accurate.
    Furthermore, even with the oracle propensity score, 
    we have just shown that the IPW has a high variance.
    On the other hand, the naive estimator (without IPW) from \S\ref{sec:naive_estimator} 
    is unbiased only when the actions in the dataset were sampled independently of the associated covariates. 
    In other words, it is often not enough to rely on either 
    of these approaches on their own
    to perform causal inference
    \citep{Belloni2011InferenceOT}.
    We can do better by 
    combining IPW with the naive estimator $\mu_A(X)$, to which we refer as a doubly robust estimator.
    In doubly robust estimation,
    a bias from one method is addressed by the other method, and vice versa \citep{Robins1994, Lunceford2004, Kang2007, Chernozhukov2017}.
    Such a method is both consistent and unbiased, as long as at least one of the IPW and the naive estimator is consistent and unbiased.
    The doubly robust estimator for ATE in the case of two actions is then
    \begin{align}
        ATE_{DR} =& \mathbb{E}_{Y,X} \left[ \mu_1(X) - \mu_0(X)\right] + \mathbb{E}_{Y,X,A}\left[A \frac{(Y - \mu_1(X))}{e(X)} -
        (1-A) \frac{(Y - \mu_0(X))}{1-e(X)}\right] 
        \label{chp2:eqn:dr}\\
        =&\mathbb{E}_{Y,X,A} \left[A \frac{(Y - \mu_1(X))}{e(X)}+\mu_1(X) \right] - \mathbb{E}_{Y,X,A} \left[(1-A) \frac{(Y - \mu_0(X))}{1-e(X)}+\mu_0(X) \right].
        \label{chp2:epn:dr}
    \end{align}
    If the estimated potential outcomes $\mu_A(X)$ are correct,
    we do not need to worry about propensity score estimation, since
    $Y - \hat{\mu}_1(X)$ and $Y - \hat{\mu}_0(X)$ will be zero in Equation~\ref{chp2:epn:dr}. 
    $\widehat{ATE}_{DR}$ hence reduces to approximating $\mathbb{E} \left[ \hat{\mu}_1(X) - \hat{\mu}_0(X)\right]$.
    In contrast, it is okay for our estimated potential outcome to be wrong if propensity score estimation is consistent. 
    If the propensity score is correctly estimated, then  $\mathbb{E}_{Y,X,A} \left[\frac{A Y}{\hat{e}(X)}\right]$ and $\mathbb{E}_{Y,X,A}\left[\frac{(1-A)Y}{(1-\hat{e}(X))}\right]$ will be weighted correctly as well, and we  
    recover the IPW estimator \citep{Robins1994, Robins1995, SCHARFSTEIN1999}.
    Consequently, such a doubly robust method is consistent 
    \citep{Hahn1998, Bang2005, Shardell2014, FARRELL20151}.
    
    Re-arranging terms and expressing in terms of Monte Carlo estimation of $ATE_{DR}$, we get
    \begin{align}
        \widehat{ATE}_{DR} \approx& 
         \frac{1}{N}\sum^{N}_{i} \left[\frac{a_iy_i}{\hat{e}(x_i)} - \frac{a_i-\hat{e}(x_i)}{\hat{e}(x_i)}\hat{\mu}_1(x_i) \right] 
        - \frac{1}{N}\sum^{N}_{i} \left[\frac{(1-a_i)y_i}{1-\hat{e}(x_i)} - \frac{\hat{e}(x_i)-a_i}{1-\hat{e}(x_i)}\hat{\mu}_0(x_i) \right] \label{chp2:epn:dr_mc1}.
    \end{align}    
    The empirical estimate converges the true ATE, $\hat{ATE}_{DR} \rightarrow ATE$, with an asymptotic variance of
    \begin{align*}
        \text{Var}[ATE_{DR}(X)] = \text{Var}[ATE_{AGG}(X)] + \mathbb{E}\left[ \frac{\sigma^2_1(X)}{e(X)} \right] + \mathbb{E}\left[ \frac{\sigma^2_0(X)}{1-e(X)} \right],
    \end{align*}
    where $\sigma_T(X) = \text{Var}[Y_i(T)|X]$.
    Despite its greater variance, the doubly robust method often exhibits greater efficiency and robustness to model misspecification. 
    
    Multiple studies have shown that doubly robust methods for ATE estimation with missing data perform better than their non-doubly robust baselines and theoretically have shown to converge faster to the true ATE than individual methods \citep{Robins1994, Imke2020}.
    A naive doubly-robust estimator is asymptotically optimal among non-parametrized estimators, meaning the semiparametric variances are bounded and asymptotically convergent if either the propensity score or the estimator $\mu_A(X)$ is correct \citep{Robins1995, Kang2007}. 
    Subsequently, other estimators with bounded asymptotic variances have been proposed even when IPW exhibits a high variance \citep{Robins2007, Tan2010BoundedEA, Waernbaum2017ModelMA}.
    


\subsection{Causal inference with representation learning} 
\label{sec:ci_representation_learning}

While matching, IPW, and doubly robust methods have their own merits for estimating potential outcomes, it may be necessary to utilize a powerful model that can express highly complex functions given high dimensional data with a limited amount of data.
For instance, working with non-linear high-dimensional data such as X-ray images may require a non-linear parametric model to transform data from its original space to a better representation that facilitates CI.
The goal is to extract a causal representation from high dimensional data (covariate and action) and more accurately predict potential outcomes from the extracted causal representation.
In this section, we review methods that extend the previous CI approaches in this way by (deep) representation learning \citep{Scholkopf2021, Wang2021}. 

Representation learning involves automatically learning features or representations of raw data. 
Rather than manual feature engineering,
representation learning enables models to learn to extract high-level features from raw, high-dimensional data such as images, audio, and texts, themselves. 
Here, the potential outcome estimator $u_A(X;\phi)$ is a deep neural network with parameter $\phi$ with $m$ 
hidden layers $H=\lbrace {\bf h}^{(i)}\rbrace^{m}_1$. Each hidden representation at $i$-th layer ($i < k$) is a function of all the previous layers, ${\bf h}_i(x) = f_i({\bf h}_{i-1}; \phi)$ with ${\bf h}_0=x$. 
Each hidden representation at $j \geq k$, ${\bf h}_j(x) = f_j({\bf h}_{j-1}, a; \phi)$, depends also on the action.

The model is trained to minimize the factual loss $l_\text{Factual}$ which is typically the negative log-likelihood of the dataset with inverse probability weighting. $u_A(X;\phi)$ learns to 
predict the potential outcome however only well on observed covariate-action pairs due to the challenges in generalization. 
We thus need to add a regularization term to the loss function in order to encourage 
the model to
generalize better to (unseen) counterfactual actions:
\begin{align}
    \mathcal{L} = \mathbb{E}_{Y,X,A}[w l_\text{Factual}(X,Y,A;u_A(X;\phi)) + \lambda \mathcal{R}({H})].
    \label{eq:dci_reg}
\end{align}
The loss function is weighted by the inverse probability weight $w=\frac{A}{2e(X)} + \frac{1-A}{2(1-e(X))}$ (see \S\ref{sec:ipw}).
Regularization often imposes certain properties on the hidden layers, which is why we refer to it as 
$\mathcal{R}(H)$, with the regularization coefficient $\lambda$~\citep{Johansson2016, Uri2017, Zhang2020, Wu2021IntactVAEET}.
Regularization reduces the hypothesis space in a way that encourages $\mu_A(X,\phi)$ to capture a causal relationship rather than a spurious relationship between the action and outcome.  
Deep representation learning based CI is vast and fast-growing \citep{Nabi2017SemiParametricCS, Yoon2018, Veitch2020a, Zhang2020b, Wang2021, Zhang2021IdentifiableER}. 
We characterize a majority of these approaches as minimizing a combination of a factual regression loss and a regularizer, as in Equation~\ref{eq:dci_reg}, and we discuss four representative ones in this section.

\begin{figure}[t]
    \begin{minipage}{0.5\linewidth}
        \centering
        \includegraphics[width=0.5\linewidth]{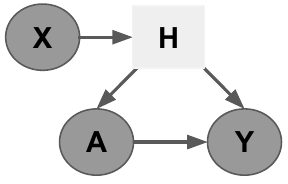}
        \subcaption{}
        \label{fig:deterministic_ci}
    \end{minipage}
    \begin{minipage}{0.5\linewidth}
        \centering
        \includegraphics[width=0.5\linewidth]{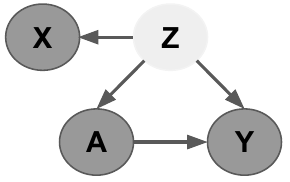}
        \subcaption{}
        \label{fig:stochastic_ci}
    \end{minipage}
    \caption{Causal Graph - (a) Deterministic Representation: $H$ corresponds to a hidden layer representation that is a deterministic variable, $X$ is a covariate, $A$ is an action, and $Y$ is an outcome variable. (b) Stochastic Representation: $Z$ is a latent variable, $X$ is a noise version of confounder $Z$, $A$ is an action, and $Y$ is an outcome variable.} 
    \label{fig:representational_learning_ci}
\end{figure}

Among these methods, we separately discuss counterfactual smoothing regularization and deep latent variable models (see Figure~\ref{fig:representational_learning_ci}) with a focus on how both of these approaches lead to better generalization of  ATE.

\subsubsection{Counterfactual Smoothing Regularization}

The objective in counterfactual smoothing is to train a model to generalize to a counterfactual potential outcome even if it only saw factual data during training (see Figure~\ref{fig:deterministic_ci}).
Although the model $u_A(X,\phi)$ is trained to estimate the potential outcome using the inverse probability weighted factual loss function, it may still underperform for an unseeded data or a data paired with unseen action (see Figure~\ref{fig:genearlization_positivity}). In such a case, the variance of the potential outcome estimate over counterfactual actions is high, implying that an individual model's prediction may be inaccurate. 

Counterfactual smoothing reduces the variance of the predicted potential outcome whose variance can be controlled by ensuring that the learned representations of treated and controlled data points to follow (largely) indistinguishable distributions in the feature space, respectively~\citep{Johansson2016, Shalit2017, Johansson2018, Yao2018}. 
The divergence between factual and counterfactual hidden representations' distributions influences the maximum variance of the ATE since covariate and an action propagate through deep neural networks, assuming that the deep neural network has a finite Lipschitz constant.
We can thus reduce the variance by optimizing the factual loss function while, for instance, minimizing the integral probability metric (IPM) as regularization. The overall objective function is
\begin{align}
    \mathcal{L}_{CFR} = \mathbb{E}_{Y,X,A}[w l_\text{Factual}(X, Y, A; u_A(X;\phi))] + \lambda \text{IPM}_{\mathcal{F}} (\hat{p}({\bf h}_i(X)|A=1), \hat{p}({\bf h}_i(X)|A=0)), 
    \label{ch2:eqn:cfr}
\end{align}
where ${\bf h}_i(X)$ is a hidden representation from the $i$-th layer of $u_A$ with $i < k$.
$\text{IPM}_{\mathcal{F}}$ is
the integral probability metric with a class $\mathcal{F}$ of real-valued bounded measurable functions.
The hidden representation ${\bf h}_i(X)$ does not depend on action $A$, although the data point $x_i$ was collected together with some action $a$. 
$\hat{p}({\bf h}_i(X)|A=1)$ and $\hat{p}({\bf h}_i(X)|A=0)$ are thus the empirical probability distributions over the representations of treated and controlled groups, respectively. The 1-Lipschitz function and universal reproducing Hilbert kernel space, such as 1-Wasserstein distance \citep{Cuturi2013FastCO} and MMD \citep{Gretton12a}, is the most commonly used class of functions for IPMs \citep{Shalit2017, Johansson2018}. 

Suppose $\text{IPM}_{\mathcal{F}} (\hat{p}({\bf h}_i(X)|A=1), \hat{p}({\bf h}_i(X))|A=0)=0$. Then, we would not be able to tell whether the hidden representation ${\bf h}_i$ is likely conditioned on action or counterfactual action. The generalization error of ITE is bounded by the CFR objective and it has been empirically shown to have lower ATE for the out of sample data. However, this does not theoretically guarantee that generalization error will be always lower. 
We explain how the approach upper bounds the PEHE estimation in Appendix~\ref{app:domain_invariance_reg}. 

\citet{Zeng2020DoubleRR} extend this approach to use a doubly robust estimator instead of IPW (see \S\ref{sec:doubly_robust}) 
and simultaneously minimize the Jensen-Shannon divergence between the treated and controlled group instead of IPM.
\citet{Hassanpour2019} view ITE estimation problem from a domain adaption perspective where factual data is assumed to come from the source and the counteraction data from a different target distribution. They use importance sampling to re-weight the loss function for each factual data to make it look like it was sampled from the target distribution instead of the source distribution \citep{Hassanpour2019}.
Instead of globally balancing the treatment and controlled posterior distributions, \citet{Yao2018} propose a local similarity preserved individual treatment effect (SITE) estimation method based on deep representation learning \citep{Yao2018}. Their method preserves the local similarity and balances between the factual and counterfactual distributions over the set of actions, simultaneously.
Furthermore, there is a line of work where they separately extract the representations of confounders and non-confounders and re-weight the confounder representation only~\citep{Kuang2017TreatmentEE, Wu2020}.

Domain invariance, or equivalently the full overlap between the factual and counterfactual distributions, can be an overly restrictive criterion, as it may remove information from input variables (covariates and sometimes action) that may be necessary for accurately estimating the treatment effect. 
\citep{Stojanov2021}. 
\citet{Zhang2020} demonstrate why it is not ideal to use a distributional divergence to balance the treated and controlled representations~\citep{Zhang2020}.
Instead, they propose to minimize the counterfactual variance and make the hidden representation invertible by adding a reconstruction loss function, 
which is, they claim, enough to have sufficient overlap in the factual and counterfactual supports.

{\em Deep kernel learning  for individual treatment effect} 
(DKLITE) is a representative passive CI method that uses variance reduction \citep{Zhang2020}.
Unlike the methods above, this algorithm manipulates the action-dependent hidden representation.
More formally, it uses kernel regression to estimate the potential outcome $\hat{y}_i = W_a {\bf h}_{m-1}(x_i) + \epsilon_{i, a}$ on top of the final layer of deep neural network ${\bf h}_{m-1}(x)$, where $W_a$ is the parameter for action $a$ and $\epsilon_{i, a}$ is an action-dependent noise variable.
The posterior distribution is $\mathcal{N}(m_a {\bf h}_{m-1}(x), \sigma^2(x;\mathcal{X},\Theta_a))$, where
$\sigma^2(x;\mathcal{X},\Theta_a)={\bf h}_{m-1}^T K^{-1}_a {\bf h}_{m-1}$ is the variance. 
DKLITE objective function minimizes both the negative log-likelihood and posterior variance given the counterfactual actions.
Although DKLITE uses kernel regression to derive the posterior distribution, it is not necessary to use kernel regression.

The objective function for training a causal inference model with posterior variance reduction can be written as
\begin{align}
    \mathcal{L}_{VR} = w \mathcal{L}_\text{Factual}(X, Y, A; u_A(X;\phi))
    + \lambda \mathbb{E}_{p(X,A)}[g(\text{VAR}[h_\theta(X,1-A))]],
    \label{ch2:eqn:vr}
\end{align}
where $\text{VAR}[h(X,1-A)]$ is the posterior variance given a data point $X=x$ and a counterfactual action.
The function $g:\mathbb{R}^D \rightarrow \mathbb{R}$ aggregates the covariance of high dimensional representations into a single scalar.
For example, $g(\cdot)$ can be a sum of the element-wise variances of the hidden representation.
By reducing the variance in the representations $h(X,1-A)$, we also reduce the variance in the ATE. 

\subsubsection{Deep Latent Variables CI Models}

In deep latent variable models for causal inference, we assume a particular data-generating process described with a probabilistic graphical model that contains latent (or hidden) stochastic variables~\citep{Louizos2017, Rakesh2018, Vowels2020, Wu2021IntactVAEET, Im2021, Kumor2021, Lu2022}.
The inclusion of such latent variables enables us to model the potential outcome with a much richer distribution. 
Without latent variables, it is challenging to build a function approximator, such as a deep neural network, that captures a multimodal distribution of the potential outcome, regardless of how complex a form such a function takes.

Causal effects are however not identifiable in general when there are latent variables. Identifiability requires that the model parameter can be uniquely estimated from the data. However, since the latent variables do not directly measure the unobserved variable but infer from the observed variables, it introduces ambiguity that leads to violates the assumption of unconfoundedness.  
In order to overcome this issue, one has to make two assumptions; (1) $X$ is a proxy variable that is a noisy version of a hidden confounder, and (2) this unknown confounder can be modelled by these latent variables (see Figure~\ref{fig:stochastic_ci}).
While these extra assumptions are a major drawback of deep latent variable-based CI~\citep{Louizos2017CausalEI, Kocaoglu2017}, we nevertheless review this literature as they are increasingly more widely used in practice \citep{Pearl2016CausalII, Wu2021IntactVAEET, Trifunov2020, Kumor2021, Rissanen2021}. 

With these assumptions, we can consider the latent variable $Z$ a hidden, i.e. unobserved, confounder, to which we have access via its noisy realization $X$.
We can recover the joint distribution $p(Z, X, Y, A)$ from observational data $(X, Y, A)$. 
We can then compute $p(Y|X=x,do(A=1))$ and $p(Y|X=x,do(A=0))$, which allows us to compute ITE, by
\begin{align*}
    p(Y|X=x,do(A=a)) &= \int_z p(Y|X,do(A=a),Z) p(Z|X=x,do(A=a))dZ\\
                     &= \int_z p(Y|X=x,do(A=a),Z) p(Z|X=x, do(A=a))dZ\\
    &= \int_z p(Y|X=x,A=a,Z) p(Z|X=x)dZ.
\end{align*}
The third equality follows from the rule of {\em do}-calculus.
Therefore, we can estimate $p(Y|X=x,do(A=a))$ as long as we can approximate $p(Y|A=a,Z)$ and $p(X|Z)$.

The causal effect variational autoencoder (CEVAE) is a particular type of variational inference framework which allows us to estimate $p(Y|A, Z)$ and $p(Z|X)$ using deep neural networks \citep{Kingma2014, Louizos2017}.
With this VAE, the true posterior distribution is defined as $p_\theta(Z|X, Y, A) \propto p_\theta(Y, A|Z)p_{\theta}(X|Z) p(Z)$, where both$p_{\theta}(Y, A |Z)$ and $p_{\theta}(X|Z)$ are modelled using a deep neural network parametrized by $\theta$ and $p(Z)$ is a prior distribution and is not parameterized by $\theta$. 
We approximate the posterior distribution with a variational posterior $q_\phi(Z|X, Y, A)$ which is modelled by a deep neural network with
a variational parameter $\phi$. We infer the hidden confounder $Z$ from the observation $(X, Y, A)$ using this variational posterior neural network.

We estimate $p(Y|A, Z)$ and $p(Z|X)$ directly by training both generative and inference networks
on observational data. 
Training is done to maximize the following variational lower bound with respect to the parameters $\theta$ and $\phi$:
\begin{align*}
    \mathcal{L}_{\text{CEVAE}} = \mathbb{E}_{p_{data}(X,Y,A)}\left[\mathbb{E}_{q_\phi(Z|X,Y,A)} \left[ \log p_\theta(X,Y,A|Z)\right] - \mathbb{KL}\left[ q_\phi(Z|X,Y,A) \| p(Z)\right]\right].
\end{align*}
The first term is the reconstruction of observable variables from the inferred confounder $Z$, and the
the second term is a regularizer which enforces the approximate posterior to be close to the prior and
maximizes the entropy of the posterior distribution over the confounder $Z$.
We jointly update both generative and inference network parameters by using backpropagation
and the re-parameterization trick \citep{Rezende2014, Kingma2014}.

Similar to CEVAE, linked causal variational autoencoder (LCVA) treats the latent attributes directly as confounders with the assumption that these confounders affect both the treatment and the outcome of units \citep{Rakesh2018}. The main difference is that the authors want to measure the causal effect when there exists {\em a spillover effect}\footnote{A spillover happens when something in one situation affects something else in a different situation, even though they may not seem related.} 
between pairs of two covariates through the confounders. 
Another variant is the Causal Effect by using Variational Information Bottleneck (CEVIB) \citep{Lu2022}. 
Just like any other variational latent model, it learns to fit the model to observation data and learns the confounders that affect treatments and outcomes using variational information bottleneck \citep{Alemi2016}.
CEVIB does this in a way that allows the model to forget some latent variables that are not confounders and
learn to extract only the confounding information from covariate.
Deep entire space cross networks for individual treatment effect estimation (DESCN) attempt to learn the
latent confounders (``the hidden treatment effect'') through a cross-network in a multi-task learning manner \citep{Zhong2022DESCNDE}. 
It reduces treatment biases, that favour one treatment over another, by learning from multiple tasks and overcoming sample imbalance.

In CEVAE, the conditional distribution $p_{\theta}(X,Y,A|Z)$ is learned from data sampled from $p(A|Z)$, $p(X|Z)$ and $p(Y|A,Z)$, where $Z\sim p(Z)$. This conditional distribution, which is used for computing treatment effect, is however used with an action $A$ sampled from $p(A)$ rather than $p(A|Z)$ in the inference time. This discrepancy is known as covariate shift or more generally distribution shift and is detrimental to generalization in deep learning \citep{Shimodaira2000, Sergey2015, Jeong2020RobustCI, Louizos2017CausalEI}, which in turn results in a degradation in the quality of ATE estimation. 


Replacing the observational distribution with a uniform treatment distribution, which is independent of the covariate, provides randomized treatment samples for training a CEVAE.
A uniform treatment selection process decouples $Z$ and $A$, thereby making $A$ independent of the covariate $X$, i.e. $p(A|X)=p(A)$. This is similar to a randomized clinical trial over treatment $A$ in Section~\ref{sec:rct}. 
For this reason, it may be beneficial to train a CEVAE using a uniform treatment distribution.
Here, the observational data-based distribution is $p(X,Y,A) = p(A|X)p(X)p(Y|X,A)$ and the corresponding uniform treatment distribution is $r(X,Y,A) = r(A)p(X)p(Y|X,A)$.

\citet{Im2021} use importance weighting to write the variational lower bound objective under the uniform treatment distribution \citep{Im2021}:
\begin{align*}
\mathcal{L}_{\text{UTVAE}} &= \mathbb{E}_{p(X,Y,A)}\left[ w(X,A) \mathbb{E}_{q_\phi(Z|X,Y,A)} \left[ \log \frac{p_\theta(X,Y,A|Z)p(Z)}{q_\phi(Z|X,Y,A)} \right] \right],
\end{align*}
where $w(X,A) = \frac{r(A|X)}{p(A|X)} = \frac{1}{2 p(A|X)} $ is the importance weight. $\frac{r(A|X)}{p(A|X)} = \frac{r(X,Y,A)}{p(X,Y,A)}$ and $r(A|X)=r(A)=\frac{1}{2}$, because of the independence between $X$ and $A$ in the causal graph and the uniformly distributed treatment selection procedure. 
UTVAE generalizes better than CEVAE especially when there is a distribution shift between training and inference.
See the details of the training procedure of CEVAE and UTVAE for generative and inference networks respectively at \citep{Im2021}.


\subsubsection{Combining active and passive methods}
\label{app:bandits_ci3}

So far, we have discussed active and passive CI learning separately. 
It is however often necessary to combine active and passive approaches in order to mitigate the bias arising from non-randomized data. 
Here, we review some of the methods that combine the two.

\citet{Sawant2018} use a bandit method to collect data online and estimate the ATE offline using the potential outcome model \citep{Sawant2018}. At each time step, they sample data using a bandit algorithm like Thompson sampling and aggregate a dataset (see Algorithm~\ref{alg1:Sawant}). Using this dataset, they update the model and re-estimate the potential outcome (see Algorithm~\ref{alg2:Sawant}). 
This approach combines active and passive learning since the model is updated in a batch training setting while the dataset can be collected asynchronously.
Similarly, \citet{Ye2020} proposes a framework that combines the two methods. Here they use  inverse probability weighting and matching algorithms for passive learning and UCB and LinUCB \citep{Slivkins2019}) for active CI algorithms \citep{Ye2020}.
In the inference time, \citet{Ye2020} estimate the potential outcomes using a passive method given a new unseen example $X=x_t$ for every action. 
If the new data point $(X=x_t, A=a_t)$ with an action $a_t$ is not in a non-randomized dataset, then the algorithm resorts to exploring the action for $X=x_t$ and adds the new data point $(X=x_t,A=a_t,Y=y_t)$ to the non-randomized dataset.


\noindent 
\begin{minipage}{0.5\linewidth}
\begin{algorithm}[H]
\caption{Online Scoring and Batch training}\label{alg1:Sawant}
\begin{algorithmic}
    \State Iteration $t=0$; Log $L=\lbrace \rbrace$; contextual distribution parameter $\theta_0$ and $\theta_1$.
    \For {$i=1,2,\cdots, T$}
        \For {$t=1,2,\cdots, T$}
        \State Sample data $x_t$
        \State Predict $\hat{Y}(1) = f(x_t,\theta_1)$
        \State Predict $\hat{Y}(0) = f(x_t,\theta_0)$
        \State Choose action $a_t = \text{argmax}_{a \in \{0, 1\}} \hat{Y}(a)$
        \State Compute $p_t = p(a_t|x_t)$
        \State $L = L \cup {(x_t,a_t, p_t)}$ 
        \EndFor
    \State Update $\theta_0$ and $\theta_1$ using Algo~\ref{alg2:Sawant}
    \EndFor
\end{algorithmic}
\end{algorithm}
\end{minipage}
\begin{minipage}{0.5\linewidth}
\begin{algorithm}[H]
\caption{Offline Batch Training}\label{alg2:Sawant}
\begin{algorithmic}
    \State Dataset $D = \lbrace \rbrace$.
    \For {$i=1,2,\cdots, T$}
        \State Sample $(x_i, a_i, p_i, y_i(a_i))$ from $L$
        \State $\hat{y}_i(a_i) = y_i(a_i) / p_i$ - bias correction
        \State $\hat{y}_i(\neg a_i) = 0$
        \For {$m=1,2,\cdots, M$}
            \State Sample $(x_m, \neg a_i, p_m, y_m(\neg a_m))$ from $L$
            \State $\hat{y}_i(\neg a_i) \leftarrow \hat{y}_i(\neg a_i) + \frac{1}{M}y_m(\neg a_i)/p_m$ 
        \EndFor
    \State $\text{CATE}_i = \hat{y}_i(a_i) - \hat{y}_i(\neg a_i)$
    \State $D = D \cup (x_i,a_i, \text{CATE}_i)$
    \EndFor
    \State update $\theta_0$ and $\theta_1$ by maximizing the likelihood on $D$.
\end{algorithmic}
\end{algorithm}
\end{minipage}
\vspace{0.5cm}

\section{Conclusion}

The objective of this paper is to introduce various algorithms and frameworks in causal inference by categorizing them into passive and active algorithms. 
We have particularly focused on estimating average treatment effect (ATE),
after outlining the standard assumptions necessary for the identification of causal effects: positivity, ignoreability, conditional exchangeability, consistency and Markov assumptions. 

We first present the randomized controlled trial (RCT) as a representative example of an active causal inference algorithm. 
We then delve into bandit approaches that aim to balance the outcome itself and the quality of estimating the treatment effect.
We explore different contextual bandit algorithms by considering various causal graph scenarios
such as taking account of non-confounding variables or dealing with unknown confounding variables.

We then move on to discussing passive CI methods, including matching, inverse probability weighting and doubly robust methods. We touched upon deep learning-based approaches as well.
A majority of these studies focus on converting a causal estimand into a statistical estimand, and subsequently,
estimating the statistical estimand in order to obtain the causal estimate.
These classical methods unfortunately do not perform well when they are put to work with high dimensional data.
In order to overcome this challenge, deep learning has been proposed as a way to learn a compact representation suitable for estimating ATE. 
We thus discussed several deep learning-based approaches that learn to 
infer causal effects by automatically extracting hidden or unknown confounders' representations in problems with high dimensional data. 

After reading the main part of this paper, readers should notice 
a clear resemblance between offline policy evaluation and passive causal inference methods. This resemblance is due to the similarity in estimating the reward in policy evaluation and estimating the potential outcomes in causal inference \citep{Swaminathan2015, li2015offline}. For example, some of the methods discussed in Section~\ref{app:bandits_ci3} have been used for offline policy evaluation in contextual bandit algorithms \citep{Li2012, Sawant2018}. 
Although we do not explore this connection further here, this is an important avenue to pursue both causal inference and reinforcement learning.

This review of causal inference is limited in two ways.
First, we do not consider collider bias, and second, we assume a stationary conditional probability distribution.
We discuss these two limitations briefly before ending the paper.

Collider bias happens when there is an extra variable, caused by both action and outcome variables, that is observed to be (conditioned on) a particular value.
For example, suppose we investigate the effect of a new versus old medication on the patient outcome, and we gather data from a hospital. 
We divide the patients into two groups: those who received both the new and old medications. 
The study finds that the patients who received the old medication had better outcomes than the new one and the hospital conclude that the old medication is more effective on patient outcomes. 
It turns out this conclusion has a collider bias.  This is because the decision to give the medication is based on the patient's recovery rate, which encourages doctors to prescribe a more potent drug. Thus, patients are more likely to receive the old medication which is stronger.  In this case, the patient's recovery rate becomes a collider variable because it is caused by both the decision to give which medication and the patient's outcome.
this is very separate from confounder bias and it doesn't address the collider bias issue 

Besides collider bias, we have assumed that all conditional distributions are stationary (i.e. do not change over the course of causal inference and data collection). 
The question is what happens if these conditional distributions shift over time?
One can introduce temporal dependencies to a causal graph to describe such shifts over time. Still, it is challenging to work with such a graph because collected data points over time are correlated with each other. 
This violates the no interference/Markov assumption we discussed earlier in Section~\ref{sec:assumption_discussion}. 
To address this problem, various time series methods have been developed that take into account the temporal dependence of the data. 
For instance, deep sequential weighting (DSW) and sequential causal effect variational autoencoder (SEVAE) estimate ITE with time-varying confounders \citep{Trifunov2022, Liu2020, Kumor2021}. 

\bibliography{main}

\begin{thebibliography}{163}
\providecommand{\natexlab}[1]{#1}
\providecommand{\url}[1]{\texttt{#1}}
\expandafter\ifx\csname urlstyle\endcsname\relax
  \providecommand{\doi}[1]{doi: #1}\else
  \providecommand{\doi}{doi: \begingroup \urlstyle{rm}\Url}\fi

\bibitem[Abadie \& Imbens(2016)Abadie and Imbens]{Abadie2016}
Alberto Abadie and Guido~W. Imbens.
\newblock Matching on the estimated propensity score.
\newblock \emph{Econometrica}, 84\penalty0 (2):\penalty0 781--807, 2016.

\bibitem[Abhijit~V. \& Esther(2012)Abhijit~V. and Esther]{Banerjee2012}
Banerjee Abhijit~V. and Duflo Esther.
\newblock Poor economics: A radical rethinking of the way to fight global poverty.
\newblock In \emph{Public Affairs}, 2012.

\bibitem[Agrawal \& Goyal(2013)Agrawal and Goyal]{Agrawal2013}
Shipra Agrawal and Navin Goyal.
\newblock Thompson sampling for contextual bandits with linear payoffs.
\newblock In \emph{Proceedings of the 30th International Conference on International Conference on Machine Learning - Volume 28}, ICML'13, pp.\  III–1220–III–1228. JMLR.org, 2013.

\bibitem[Alemi et~al.(2016)Alemi, Fischer, Dillon, and Murphy]{Alemi2016}
Alexander~A. Alemi, Ian Fischer, Joshua~V. Dillon, and Kevin Murphy.
\newblock Deep variational information bottleneck.
\newblock 2016.

\bibitem[Amusa(2018)]{Amusa2018}
Lateef Amusa.
\newblock Reducing bias in observational studies: An empirical comparison of propensity score matching methods.
\newblock \emph{Turkiye Klinikleri Journal of Biostatistics}, 10:\penalty0 13--26, 01 2018.
\newblock \doi{10.5336/biostatic.2017-58633}.

\bibitem[Athey et~al.(2015)Athey, Imbens, and Ramachandra]{Athey2015}
Susan Athey, Guido Imbens, and Vikas Ramachandra.
\newblock Machine learning methods for estimating heterogeneous causal effects.
\newblock 04 2015.

\bibitem[Auer et~al.("2002")Auer, Fischer, and Cesa-Bianchi"]{Auer2002b}
"Peter Auer, Paul Fischer, and Nicolo Cesa-Bianchi".
\newblock "finite-time analysis of the multiarmed bandit problem".
\newblock \emph{"Machine Learning"}, "47"\penalty0 ("3"):\penalty0 "235--256", "2002".

\bibitem[Bareinboim \& Pearl(2016)Bareinboim and Pearl]{Bareinboim2016}
Elias Bareinboim and Judea Pearl.
\newblock Causal inference and the data-fusion problem.
\newblock \emph{Proceedings of the National Academy of Sciences}, 113\penalty0 (27):\penalty0 7345--7352, 2016.
\newblock \doi{10.1073/pnas.1510507113}.
\newblock URL \url{https://www.pnas.org/doi/abs/10.1073/pnas.1510507113}.

\bibitem[Bareinboim et~al.(2015)Bareinboim, Forney, and Pearl]{Bareinboim2015}
Elias Bareinboim, Andrew Forney, and Judea Pearl.
\newblock Bandits with unobserved confounders: A causal approach.
\newblock In C.~Cortes, N.~Lawrence, D.~Lee, M.~Sugiyama, and R.~Garnett (eds.), \emph{Advances in Neural Information Processing Systems}, volume~28. Curran Associates, Inc., 2015.
\newblock URL \url{https://proceedings.neurips.cc/paper/2015/file/795c7a7a5ec6b460ec00c5841019b9e9-Paper.pdf}.

\bibitem[Belloni et~al.(2011)Belloni, Chernozhukov, and Hansen]{Belloni2011InferenceOT}
Alexandre Belloni, Victor Chernozhukov, and Christian Hansen.
\newblock Inference on treatment effects after selection amongst high-dimensional controls.
\newblock \emph{Operations Research eJournal}, 2011.

\bibitem[Bender(2020)]{Bender2020}
Andrea Bender.
\newblock What is causal cognition?
\newblock \emph{Frontiers in Psychology}, 11, 01 2020.
\newblock \doi{10.3389/fpsyg.2020.00003}.

\bibitem[Berry \& Fristedt(1985)Berry and Fristedt]{berry1985bandit}
Donald~A Berry and Bert Fristedt.
\newblock Bandit problems: sequential allocation of experiments (monographs on statistics and applied probability).
\newblock \emph{London: Chapman and Hall}, 5\penalty0 (71-87):\penalty0 7--7, 1985.

\bibitem[Bouneffouf et~al.(2020)Bouneffouf, Rish, and Aggarwal]{Bouneffouf2020}
Djallel Bouneffouf, Irina Rish, and Charu Aggarwal.
\newblock Survey on applications of multi-armed and contextual bandits.
\newblock In \emph{2020 IEEE Congress on Evolutionary Computation (CEC)}, pp.\  1--8, 2020.
\newblock \doi{10.1109/CEC48606.2020.9185782}.

\bibitem[Breslow et~al.(2009)Breslow, Lumley, Ballantyne, Chambless, and Kulich]{Breslow2009}
Norman~E. Breslow, Thomas Lumley, Christie~M. Ballantyne, Lloyd~E. Chambless, and Michal Kulich.
\newblock Using the whole cohort in the analysis of case-cohort data.
\newblock \emph{American journal of epidemiology}, 169:\penalty0 1398--1405, 2009.

\bibitem[Carlson(2019)]{Carlson2019}
Bruce~W. Carlson.
\newblock Simpson’s paradox.
\newblock \emph{Encyclopedia Britannica}, 2019.
\newblock \doi{https://www.britannica.com/topic/Simpsons-paradox}.

\bibitem[Castro et~al.(2020)Castro, Walker, and Glocker]{Castro2020}
Daniel~C. Castro, Ian Walker, and Ben Glocker.
\newblock Causality matters in medical imaging.
\newblock \emph{Nature Communication}, 3673, 2020.
\newblock \doi{https://doi.org/10.1038/s41467-020-17478-w}.

\bibitem[Chalmers et~al.(1981)Chalmers, Smith, Blackburn, Silverman, Schroeder, Reitman, and Ambroz]{Chalmers1981}
T.~C. Chalmers, Jr~Smith, H., B.~Blackburn, B.~Silverman, B.~Schroeder, D.~Reitman, and A.~Ambroz.
\newblock A method for assessing the quality of a randomized control trial.
\newblock \emph{Controlled clinical trials}, 2:\penalty0 31--49, 1981.

\bibitem[Chalmers et~al.(1989)Chalmers, Hewett, Reitman, and Sacks]{Chalmers1989SelectionAE}
Thomas~C. Chalmers, Paul Hewett, Dinah Reitman, and Henry~S Sacks.
\newblock Selection and evaluation of empirical research in technology assessment.
\newblock \emph{International Journal of Technology Assessment in Health Care}, 5:\penalty0 521 -- 536, 1989.

\bibitem[Chapelle et~al.(2015)Chapelle, Manavoglu, and Rosales]{Chapelle2015}
Olivier Chapelle, Eren Manavoglu, and Romer Rosales.
\newblock Simple and scalable response prediction for display advertising.
\newblock \emph{ACM Transaction Intelligence System Technology}, 5\penalty0 (4), 2015.

\bibitem[Chernozhukov et~al.(2017)Chernozhukov, Chetverikov, Demirer, Duflo, Hansen, Newey, and Robins]{Chernozhukov2017}
Victor Chernozhukov, Denis Chetverikov, Mert Demirer, Esther Duflo, Christian Hansen, Whitney Newey, and James~M. Robins.
\newblock Double/debiased machine learning for treatment and structural parameters.
\newblock \emph{Econometrics: Econometric \& Statistical Methods - Special Topics eJournal}, 2017.

\bibitem[Choudhry(2017)]{Choudhry2017RandomizedCT}
Niteesh~K. Choudhry.
\newblock Randomized, controlled trials in health insurance systems.
\newblock \emph{The New England Journal of Medicine}, 377:\penalty0 957–964, 2017.

\bibitem[Cochran \& Rubin(1973)Cochran and Rubin]{Cochran1973}
William~G. Cochran and Donald~B. Rubin.
\newblock Controlling bias in observational studies: A review.
\newblock \emph{The Indian Journal of Statistics, Series A}, 35:\penalty0 417--446, 1973.
\newblock \doi{JSTOR, http://www.jstor.org/stable/25049893}.

\bibitem[Concato et~al.(2000)Concato, Shah, and Horwitz]{Concato2000}
John Concato, Nirav Shah, and Ralph~I. Horwitz.
\newblock Randomized, controlled trials, observational studies, and the hierarchy of research designs.
\newblock \emph{New England Journal of Medicine}, 342\penalty0 (25):\penalty0 1887--1892, 2000.

\bibitem[Cuturi \& Doucet(2013)Cuturi and Doucet]{Cuturi2013FastCO}
Marco Cuturi and A.~Doucet.
\newblock Fast computation of wasserstein barycenters.
\newblock In \emph{International Conference on Machine Learning}, 2013.

\bibitem[Danilo~Jimenez \& Shakir(2014)Danilo~Jimenez and Shakir]{Rezende2014}
Rezende Danilo~Jimenez and Mohamed Shakir.
\newblock Stochastic backpropagation and approximate inference in deep generative models.
\newblock In \emph{arXiv preprint arXiv:1401.4082}, 2014.

\bibitem[Deaton \& Cartwright(2016)Deaton and Cartwright]{Deaton2016UnderstandingAM}
Angus Deaton and Nancy Cartwright.
\newblock Understanding and misunderstanding randomized controlled trials.
\newblock \emph{Behavioral \& Experimental Economics eJournal}, 2016.

\bibitem[Dehejia \& Wahba(2002)Dehejia and Wahba]{Dehejia2002}
Rajeev Dehejia and Sadek Wahba.
\newblock Propensity score matching methods for non-experimental causal studies.
\newblock \emph{The Review of Economics and Statistics}, 84:\penalty0 151--161, 02 2002.
\newblock \doi{10.1162/003465302317331982}.

\bibitem[Dehejia \& Wahba(1999)Dehejia and Wahba]{Dehejia1999}
Rajeev~H. Dehejia and Sadek Wahba.
\newblock Causal effects in nonexperimental studies: Reevaluating the evaluation of training programs.
\newblock \emph{Journal of the American Statistical Association}, 94\penalty0 (448):\penalty0 1053--1062, 1999.
\newblock \doi{10.1080/01621459.1999.10473858}.
\newblock URL \url{https://www.tandfonline.com/doi/abs/10.1080/01621459.1999.10473858}.

\bibitem[Diamond \& Sekhon(2013)Diamond and Sekhon]{Diamond2013}
Alexis Diamond and Jasjeet~S. Sekhon.
\newblock {Genetic Matching for Estimating Causal Effects: A General Multivariate Matching Method for Achieving Balance in Observational Studies}.
\newblock \emph{The Review of Economics and Statistics}, 95\penalty0 (3):\penalty0 932--945, 2013.

\bibitem[Dimakopoulou et~al.(2017)Dimakopoulou, Athey, and Imbens]{Dimakopoulou2017}
Maria Dimakopoulou, Susan Athey, and Guido Imbens.
\newblock Estimation considerations in contextual bandits.
\newblock 11 2017.

\bibitem[Farrell(2015)]{FARRELL20151}
Max~H. Farrell.
\newblock Robust inference on average treatment effects with possibly more covariates than observations.
\newblock \emph{Journal of Econometrics}, 189\penalty0 (1):\penalty0 1--23, 2015.
\newblock ISSN 0304-4076.
\newblock \doi{https://doi.org/10.1016/j.jeconom.2015.06.017}.
\newblock URL \url{https://www.sciencedirect.com/science/article/pii/S0304407615001864}.

\bibitem[Ferraro et~al.(2019)Ferraro, Sanchirico, and Smith]{ferraro2019}
Paul~J. Ferraro, James~N. Sanchirico, and Martin~D. Smith.
\newblock Causal inference in coupled human and natural systems.
\newblock \emph{Proceedings of the National Academy of Sciences}, 116\penalty0 (12):\penalty0 5311--5318, 2019.
\newblock \doi{10.1073/pnas.1805563115}.
\newblock URL \url{https://www.pnas.org/doi/abs/10.1073/pnas.1805563115}.

\bibitem[Frieden(2017)]{Frieden2017EvidenceFH}
Thomas~R. Frieden.
\newblock Evidence for health decision making — beyond randomized, controlled trials: The changing face of clinical trials.
\newblock \emph{The New England Journal of Medicine}, 377:\penalty0 465–475, 2017.

\bibitem[Féraud et~al.(2016)Féraud, Allesiardo, Urvoy, and Clérot]{Feraud2016}
Raphaël Féraud, Robin Allesiardo, Tanguy Urvoy, and Fabrice Clérot.
\newblock Random forest for the contextual bandit problem.
\newblock In Arthur Gretton and Christian~C. Robert (eds.), \emph{Proceedings of the 19th International Conference on Artificial Intelligence and Statistics}, volume~51 of \emph{Proceedings of Machine Learning Research}, pp.\  93--101, Cadiz, Spain, 09--11 May 2016. PMLR.

\bibitem[Garc{\'i}a \& Wantchekon(2010)Garc{\'i}a and Wantchekon]{MartelGarca2010TheoryEV}
Fernando~Martel Garc{\'i}a and L{\'e}onard Wantchekon.
\newblock Theory, external validity, and experimental inference: Some conjectures.
\newblock \emph{The ANNALS of the American Academy of Political and Social Science}, 628:\penalty0 132 -- 147, 2010.

\bibitem[Gordon et~al.(2019)Gordon, Zettelmeyer, Bhargava, and Chapsky]{Gordon2019}
Brett~R. Gordon, Florian Zettelmeyer, Neha Bhargava, and Dan Chapsky.
\newblock {A Comparison of Approaches to Advertising Measurement: Evidence from Big Field Experiments at Facebook}.
\newblock \emph{Marketing Science}, 38\penalty0 (2):\penalty0 193--225, 2019.

\bibitem[Graepel et~al.(2010)Graepel, Quiñonero~Candela, Borchert, and Herbrich]{Graepel2010}
Thore Graepel, Joaquin Quiñonero~Candela, Thomas Borchert, and Ralf Herbrich.
\newblock Web-scale bayesian click-through rate prediction for sponsored search advertising in microsoft's bing search engine.
\newblock In \emph{Proceedings of the 27th International Conference on Machine Learning ICML 2010, Invited Applications Track (unreviewed, to appear)}, June 2010.
\newblock Invited Applications Track.

\bibitem[Green \& Glasgow(2006)Green and Glasgow]{Green2006EvaluatingTR}
L.~W. Green and Russell~E. Glasgow.
\newblock Evaluating the relevance, generalization, and applicability of research.
\newblock \emph{Evaluation \& the Health Professions}, 29:\penalty0 126 -- 153, 2006.

\bibitem[Greenewald et~al.(2017)Greenewald, Tewari, Murphy, and Klasnja]{Greenewald2017}
Kristjan Greenewald, Ambuj Tewari, Susan Murphy, and Predag Klasnja.
\newblock Action centered contextual bandits.
\newblock In I.~Guyon, U.~Von Luxburg, S.~Bengio, H.~Wallach, R.~Fergus, S.~Vishwanathan, and R.~Garnett (eds.), \emph{Advances in Neural Information Processing Systems}, volume~30. Curran Associates, Inc., 2017.
\newblock URL \url{https://proceedings.neurips.cc/paper/2017/file/4fa177df22864518b2d7818d4db5db2d-Paper.pdf}.

\bibitem[Gretton et~al.(2012)Gretton, Borgwardt, Rasch, Sch{{\"o}}lkopf, and Smola]{Gretton12a}
Arthur Gretton, Karsten~M. Borgwardt, Malte~J. Rasch, Bernhard Sch{{\"o}}lkopf, and Alexander Smola.
\newblock A kernel two-sample test.
\newblock \emph{Journal of Machine Learning Research}, 13\penalty0 (25):\penalty0 723--773, 2012.
\newblock URL \url{http://jmlr.org/papers/v13/gretton12a.html}.

\bibitem[Gu \& Rosenbaum(1993)Gu and Rosenbaum]{Gu1993ComparisonOM}
Xing Gu and Paul~R. Rosenbaum.
\newblock Comparison of multivariate matching methods: Structures, distances, and algorithms.
\newblock \emph{Journal of Computational and Graphical Statistics}, 2:\penalty0 405--420, 1993.

\bibitem[Hahn(1998)]{Hahn1998}
Jinyong Hahn.
\newblock {On the Role of the Propensity Score in Efficient Semiparametric Estimation of Average Treatment Effects}.
\newblock \emph{Econometrica}, 66\penalty0 (2):\penalty0 315--332, March 1998.

\bibitem[Hannan(2008)]{Hannan2008}
Edward~L Hannan.
\newblock Randomized clinical trials and observational studies: guidelines for assessing respective strengths and limitations.
\newblock \emph{JACC. Cardiovascular interventions}, 1:\penalty0 211--217, 2008.

\bibitem[Hansen(2004)]{Hansen2004}
Ben Hansen.
\newblock Full matching in an observational study of coaching for the sat.
\newblock \emph{Journal of the American Statistical Association}, 99:\penalty0 609--618, 02 2004.
\newblock \doi{10.1198/016214504000000647}.

\bibitem[Hassanpour \& Greiner(2019)Hassanpour and Greiner]{Hassanpour2019}
Negar Hassanpour and Russell Greiner.
\newblock Counterfactual regression with importance sampling weights.
\newblock In \emph{International Joint Conference on Artificial Intelligence}, 2019.

\bibitem[Heckman et~al.(1997)Heckman, Ichimura, and Todd]{Heckman1997}
James Heckman, Hidehiko Ichimura, and Petra Todd.
\newblock Matching as an econometric evaluation estimator: Evidence from evaluating a job training programme.
\newblock \emph{Review of Economic Studies}, 64:\penalty0 605--54, 02 1997.
\newblock \doi{10.2307/2971733}.

\bibitem[Heckman \& Robb(1985)Heckman and Robb]{HECKMAN1985239}
James~J. Heckman and Richard Robb.
\newblock Alternative methods for evaluating the impact of interventions: An overview.
\newblock \emph{Journal of Econometrics}, 30\penalty0 (1):\penalty0 239--267, 1985.
\newblock ISSN 0304-4076.
\newblock \doi{https://doi.org/10.1016/0304-4076(85)90139-3}.
\newblock URL \url{https://www.sciencedirect.com/science/article/pii/0304407685901393}.

\bibitem[Heejung \& James~M.(2005)Heejung and James~M.]{Bang2005}
Bang Heejung and Robins James~M.
\newblock {Doubly robust estimation in missing data and causal inference models}.
\newblock \emph{Biometrics}, 61\penalty0 (4):\penalty0 962--973, 07 2005.
\newblock \doi{10.1111/j.1541-0420.2005.00377.x}.

\bibitem[Hernán \& Robins(2006)Hernán and Robins]{Hernan2006}
Miguel~A Hernán and James~M Robins.
\newblock Estimating causal effects from epidemiological data.
\newblock \emph{Journal of epidemiology and community health}, 60:\penalty0 578--586, 2006.

\bibitem[Hernán et~al.(2019)Hernán, Hsu, and Healy]{Miguel2019}
Miguel~A. Hernán, John Hsu, and Brian Healy.
\newblock A second chance to get causal inference right: A classification of data science tasks.
\newblock \emph{CHANCE}, 32\penalty0 (1):\penalty0 42--49, 2019.
\newblock \doi{10.1080/09332480.2019.1579578}.
\newblock URL \url{https://doi.org/10.1080/09332480.2019.1579578}.

\bibitem[Hirano et~al.(2003)Hirano, Imbens, and Ridder]{Hirano2003}
Keisuke Hirano, Guido Imbens, and Geert Ridder.
\newblock Efficient estimation of average treatment effects using the estimated propensity score.
\newblock \emph{Econometrica}, 71:\penalty0 1161--1189, 02 2003.
\newblock \doi{10.1111/1468-0262.00442}.

\bibitem[Holland(1986)]{Holland1986}
Paul~W. Holland.
\newblock Statistics and causal inference.
\newblock \emph{Journal of the American Statistical Association}, 81\penalty0 (396):\penalty0 945--960, 1986.
\newblock \doi{10.1080/01621459.1986.10478354}.
\newblock URL \url{https://www.tandfonline.com/doi/abs/10.1080/01621459.1986.10478354}.

\bibitem[Iacus et~al.(2012)Iacus, King, and Porro]{Stefano2012}
Stefano~M. Iacus, Gary King, and Giuseppe Porro.
\newblock Causal inference without balance checking: Coarsened exact matching.
\newblock \emph{Political Analysis}, 20\penalty0 (1):\penalty0 1–24, 2012.
\newblock \doi{10.1093/pan/mpr013}.

\bibitem[Imbens(2004)]{imbens2004nonparametric}
Guido~W Imbens.
\newblock Nonparametric estimation of average treatment effects under exogeneity: A review.
\newblock \emph{Review of Economics and statistics}, 86\penalty0 (1):\penalty0 4--29, 2004.

\bibitem[Imbens \& Rubin(2015)Imbens and Rubin]{Imbens2015}
Guido~W. Imbens and Donald~B. Rubin.
\newblock Causal inference for statistics, social, and biomedical sciences.
\newblock \emph{Cambridge University Press}, 2015.

\bibitem[Jalan \& Ravallion(2001)Jalan and Ravallion]{Jalan2001}
Jyotsna Jalan and Martin Ravallion.
\newblock Estimating the benefit incidence of an antipoverty program by propensity-score matching.
\newblock \emph{Journal of Business and Economic Statistics}, 21, 12 2001.
\newblock \doi{10.1198/073500102288618720}.

\bibitem[Jane-wit et~al.(2010)Jane-wit, Horwitz, and Concato]{Janewit2010}
Dan Jane-wit, Ralph~I Horwitz, and John Concato.
\newblock Variation in results from randomized, controlled trials: stochastic or systematic?
\newblock \emph{Journal of clinical epidemiology}, 63 1:\penalty0 56--63, 2010.

\bibitem[Jeong \& Namkoong(2020)Jeong and Namkoong]{Jeong2020RobustCI}
Sookyo Jeong and Hongseok Namkoong.
\newblock Robust causal inference under covariate shift via worst-case subpopulation treatment effects.
\newblock \emph{ArXiv}, abs/2007.02411, 2020.

\bibitem[jiwoong Im et~al.(2021)jiwoong Im, Cho, and Razavian]{Im2021}
Daniel jiwoong Im, Kyunghyun Cho, and Narges Razavian.
\newblock Causal effect variational autoencoder with uniform treatment for overcoming covariate shifts.
\newblock In \emph{arXiv preprint arXiv:2111.08656}, 2021.

\bibitem[Johansson et~al.(2016)Johansson, Shalit, and Sontag]{Johansson2016}
Fredrik~D. Johansson, Uri Shalit, and David Sontag.
\newblock Learning representations for counterfactual inference.
\newblock In \emph{arXiv preprint arXiv:1605.03661}, 2016.

\bibitem[Johansson et~al.(2018)Johansson, Kallus, Shalit, and Sontag]{Johansson2018}
Fredrik~D. Johansson, Nathan Kallus, Uri Shalit, and David~A. Sontag.
\newblock Learning weighted representations for generalization across designs.
\newblock \emph{arXiv: Machine Learning}, 2018.

\bibitem[Kang \& Schafer(2007)Kang and Schafer]{Kang2007}
Joseph Kang and Joseph~L. Schafer.
\newblock Demystifying double robustness: A comparison of alternative strategies for estimating a population mean from incomplete data.
\newblock \emph{Statistical Science}, 22:\penalty0 523--539, 2007.

\bibitem[Karp(1972)]{Karp1972}
Richard Karp.
\newblock Reducibility among combinatorial problems.
\newblock volume~40, pp.\  85--103, 01 1972.
\newblock ISBN 978-3-540-68274-5.
\newblock \doi{10.1007/978-3-540-68279-0-8}.

\bibitem[Kendall(2003)]{Kendall12003}
J~M Kendall.
\newblock Designing a research project: randomised controlled trials and their principles.
\newblock \emph{Emergency Medicine Journal}, 20\penalty0 (2):\penalty0 164--168, 2003.
\newblock ISSN 1472-0205.
\newblock \doi{10.1136/emj.20.2.164}.
\newblock URL \url{https://emj.bmj.com/content/20/2/164}.

\bibitem[Kidd \& Hayden(2015)Kidd and Hayden]{Celeste2015}
Celeste Kidd and {Benjamin Y.} Hayden.
\newblock The psychology and neuroscience of curiosity.
\newblock \emph{Neuron}, 88\penalty0 (3):\penalty0 449--460, November 2015.
\newblock ISSN 0896-6273.
\newblock \doi{10.1016/j.neuron.2015.09.010}.
\newblock Funding Information: This research was supported by a grant from the NIH, R01 (DA038615) (to B.Y.H.). We thank Sarah Heilbronner, Steve Piantadosi, Shraddha Shah, Maya Wang, Habiba Azab, and Maddie Pelz for helpful comments. Publisher Copyright: {\textcopyright} 2015 Elsevier Inc.

\bibitem[Kim \& Steiner(2016)Kim and Steiner]{Kim2016}
Yongnam Kim and Peter Steiner.
\newblock Quasi-experimental designs for causal inference.
\newblock \emph{Educ Psychol}, 51:\penalty0 395--405, 2016.
\newblock \doi{10.1080/00461520.2016.1207177}.

\bibitem[Kingma \& Welling(2014)Kingma and Welling]{Kingma2014}
Diederik~P. Kingma and Max Welling.
\newblock Auto-encoding variational bayes.
\newblock In \emph{arXiv preprint arXiv:1312.6114}, 2014.

\bibitem[Kocaoglu et~al.(2017)Kocaoglu, Shanmugam, and Bareinboim]{Kocaoglu2017}
Murat Kocaoglu, Karthikeyan Shanmugam, and Elias Bareinboim.
\newblock Experimental design for learning causal graphs with latent variables.
\newblock In I.~Guyon, U.~Von Luxburg, S.~Bengio, H.~Wallach, R.~Fergus, S.~Vishwanathan, and R.~Garnett (eds.), \emph{Advances in Neural Information Processing Systems}, volume~30. Curran Associates, Inc., 2017.
\newblock URL \url{https://proceedings.neurips.cc/paper/2017/file/291d43c696d8c3704cdbe0a72ade5f6c-Paper.pdf}.

\bibitem[Krishnamurthy et~al.(2018)Krishnamurthy, Wu, and Syrgkanis]{Krishnamurthy2018}
Akshay Krishnamurthy, Zhiwei~Steven Wu, and Vasilis Syrgkanis.
\newblock Semiparametric contextual bandits.
\newblock arXiv, 2018.
\newblock \doi{10.48550/ARXIV.1803.04204}.
\newblock URL \url{https://arxiv.org/abs/1803.04204}.

\bibitem[Kuang et~al.(2017)Kuang, Cui, Li, Jiang, Yang, and Wang]{Kuang2017TreatmentEE}
Kun Kuang, Peng Cui, B.~Li, Meng Jiang, Shiqiang Yang, and Fei Wang.
\newblock Treatment effect estimation with data-driven variable decomposition.
\newblock In \emph{AAAI Conference on Artificial Intelligence}, 2017.

\bibitem[Kumor et~al.(2021)Kumor, Zhang, and Bareinboim]{Kumor2021}
Daniel Kumor, Junzhe Zhang, and Elias Bareinboim.
\newblock Sequential causal imitation learning with unobserved confounders.
\newblock In M.~Ranzato, A.~Beygelzimer, Y.~Dauphin, P.S. Liang, and J.~Wortman Vaughan (eds.), \emph{Advances in Neural Information Processing Systems}, volume~34, pp.\  14669--14680. Curran Associates, Inc., 2021.
\newblock URL \url{https://proceedings.neurips.cc/paper/2021/file/7b670d553471ad0fd7491c75bad587ff-Paper.pdf}.

\bibitem[LaLonde(1986)]{LaLonde1986}
Robert~J. LaLonde.
\newblock Evaluating the econometric evaluations of training programs with experimental data.
\newblock \emph{The American economic review}, pp.\  604--620, 1986.

\bibitem[Langford \& Zhang(2007)Langford and Zhang]{Langford2007}
John Langford and Tong Zhang.
\newblock The epoch-greedy algorithm for multi-armed bandits with side information.
\newblock In J.~Platt, D.~Koller, Y.~Singer, and S.~Roweis (eds.), \emph{Advances in Neural Information Processing Systems}, volume~20. Curran Associates, Inc., 2007.
\newblock URL \url{https://proceedings.neurips.cc/paper/2007/file/4b04a686b0ad13dce35fa99fa4161c65-Paper.pdf}.

\bibitem[Lattimore et~al.(2016)Lattimore, Lattimore, and Reid]{Lattimore2016}
Finnian Lattimore, Tor Lattimore, and Mark~D. Reid.
\newblock Causal bandits: Learning good interventions via causal inference, 2016.
\newblock URL \url{https://arxiv.org/abs/1606.03203}.

\bibitem[Li(2015)]{li2015offline}
Lihong Li.
\newblock Offline evaluation and optimization for interactive systems.
\newblock In \emph{Proceedings of the 8th ACM International Conference on Web Search and Data Mining}. ACM - Association for Computing Machinery, February 2015.
\newblock URL \url{https://www.microsoft.com/en-us/research/publication/offline-evaluation-and-optimization-for-interactive-systems/}.

\bibitem[Li et~al.(2010)Li, Chu, Langford, and Schapire]{Li2010}
Lihong Li, Wei Chu, John Langford, and Robert~E. Schapire.
\newblock A contextual-bandit approach to personalized news article recommendation.
\newblock In \emph{Proceedings of the 19th International Conference on World Wide Web}, WWW '10, pp.\  661–670, New York, NY, USA, 2010. Association for Computing Machinery.
\newblock ISBN 9781605587998.
\newblock \doi{10.1145/1772690.1772758}.
\newblock URL \url{https://doi.org/10.1145/1772690.1772758}.

\bibitem[Li et~al.(2012)Li, Chu, Langford, Moon, and Wang]{Li2012}
Lihong Li, Wei Chu, John Langford, Taesup Moon, and Xuanhui Wang.
\newblock An unbiased offline evaluation of contextual bandit algorithms with generalized linear models.
\newblock In Dorota Glowacka, Louis Dorard, and John Shawe-Taylor (eds.), \emph{Proceedings of the Workshop on On-line Trading of Exploration and Exploitation 2}, volume~26 of \emph{Proceedings of Machine Learning Research}, pp.\  19--36, Bellevue, Washington, USA, 02 Jul 2012. PMLR.

\bibitem[Liu et~al.(2020)Liu, Yin, and Zhang]{Liu2020}
Ruoqi Liu, Changchang Yin, and Ping Zhang.
\newblock Estimating individual treatment effects with time-varying confounders, 2020.

\bibitem[Louizos et~al.(2017{\natexlab{a}})Louizos, Shalit, Mooij, Sontag, Zemel, and Welling]{Louizos2017}
Christos Louizos, Uri Shalit, Joris Mooij, David Sontag, Richard Zemel, and Max Welling.
\newblock Causal effect inference with deep latent-variable models.
\newblock In \emph{arXiv preprint arXiv:1705.08821}, 2017{\natexlab{a}}.

\bibitem[Louizos et~al.(2017{\natexlab{b}})Louizos, Shalit, Mooij, Sontag, Zemel, and Welling]{Louizos2017CausalEI}
Christos Louizos, Uri Shalit, Joris~M. Mooij, David~A. Sontag, Richard~S. Zemel, and Max Welling.
\newblock Causal effect inference with deep latent-variable models.
\newblock In \emph{NIPS}, 2017{\natexlab{b}}.

\bibitem[Lu \& Rosenbaum(2004{\natexlab{a}})Lu and Rosenbaum]{Bo2004}
Bo~Lu and Paul Rosenbaum.
\newblock Optimal pair matching with two control groups.
\newblock \emph{Journal of Computational and Graphical Statistics - J COMPUT GRAPH STAT}, 13:\penalty0 422--434, 06 2004{\natexlab{a}}.
\newblock \doi{10.1198/1061860043470}.

\bibitem[Lu \& Rosenbaum(2004{\natexlab{b}})Lu and Rosenbaum]{Lu2004OptimalPM}
Bo~Lu and Paul~R. Rosenbaum.
\newblock Optimal pair matching with two control groups.
\newblock \emph{Journal of Computational and Graphical Statistics}, 13:\penalty0 422 -- 434, 2004{\natexlab{b}}.

\bibitem[Lu et~al.(2022)Lu, Cheng, Zhong, Stoian, Yuan, and Wang]{Lu2022}
Zhenyu Lu, Yurong Cheng, Mingjun Zhong, George Stoian, Ye~Yuan, and Guoren Wang.
\newblock \emph{Causal Effect Estimation Using Variational Information Bottleneck}, pp.\  288--296.
\newblock 12 2022.
\newblock ISBN 978-3-031-20308-4.
\newblock \doi{10.1007/978-3-031-20309-1-25}.

\bibitem[Lunceford \& Davidian(2004)Lunceford and Davidian]{Lunceford2004}
Jared~K Lunceford and Marie Davidian.
\newblock Stratification and weighting via the propensity score in estimation of causal treatment effects: a comparative study.
\newblock \emph{Statistics in Medicine}, 23, 2004.

\bibitem[Mayer et~al.(2020)Mayer, Sverdrup, Gauss, Moyer, Wager, and Josse]{Imke2020}
Imke Mayer, Erik Sverdrup, Tobias Gauss, Jean-Denis Moyer, Stefan Wager, and Julie Josse.
\newblock {Doubly robust treatment effect estimation with missing attributes}.
\newblock \emph{The Annals of Applied Statistics}, 14\penalty0 (3):\penalty0 1409 -- 1431, 2020.
\newblock \doi{10.1214/20-AOAS1356}.
\newblock URL \url{https://doi.org/10.1214/20-AOAS1356}.

\bibitem[Monahan et~al.(2011)Monahan, Williams, and Steinberg]{Monahan2011}
Kathryn Monahan, Joanna Williams, and Laurence Steinberg.
\newblock Revisiting the impact of part-time work on adolescent adjustment: Distinguishing between selection and socialization using propensity score matching.
\newblock \emph{Child development}, 82:\penalty0 96--112, 01 2011.
\newblock \doi{10.1111/j.1467-8624.2010.01543.x}.

\bibitem[Moreira \& Susser(2002)Moreira and Susser]{Moreira2002}
Edson~Duarte Moreira and Ezra~S Susser.
\newblock Guidelines on how to assess the validity of results presented in subgroup analysis of clinical trials.
\newblock \emph{Revista do Hospital das Clinicas}, 57 2:\penalty0 83--8, 2002.

\bibitem[Musci \& Stuart(2019)Musci and Stuart]{Rashelle2019}
Rashelle~J. Musci and Elizabeth~A. Stuart.
\newblock Ensuring causal, not casual, inference.
\newblock \emph{Prevention Science}, 20:\penalty0 452--456, 2019.

\bibitem[Nabi \& Shpitser(2017)Nabi and Shpitser]{Nabi2017SemiParametricCS}
Razieh Nabi and Ilya Shpitser.
\newblock Semi-parametric causal sufficient dimension reduction of high dimensional treatments.
\newblock \emph{arXiv: Methodology}, 2017.

\bibitem[Olivo et~al.(2008)Olivo, Macedo, Gadotti, Fuentes, Stanton, and Magee]{Olivo2008ScalesTA}
Susan~Armijo Olivo, Luciana~Gazzi Macedo, Inae~Caroline Gadotti, Jorge Fuentes, Tasha~R. Stanton, and D~J Magee.
\newblock Scales to assess the quality of randomized controlled trials: A systematic review.
\newblock \emph{Physical Therapy}, 88:\penalty0 156 -- 175, 2008.

\bibitem[P.~M. et~al.(2021)P.~M., James~M., Theo, Fredrik, and Jasjeet]{Aronow2021}
Aronow P.~M., Robins James~M., Saarinen Theo, Sävje Fredrik, and Sekhon Jasjeet.
\newblock Nonparametric identification is not enough, but randomized controlled trials are.
\newblock In \emph{arXiv preprint arXiv:2108.11342}, 2021.

\bibitem[Pearl(2009)]{Pearl09a}
Judea Pearl.
\newblock \emph{Causality: Models, Reasoning and Inference}.
\newblock Cambridge University Press, 2nd edition, 2009.

\bibitem[Pearl(2010)]{Pearl2010}
Judea Pearl.
\newblock An introduction to causal inference.
\newblock \emph{The international journal of biostatistics}, 6:\penalty0 1557--4679, 2010.

\bibitem[Pearl(2015)]{Pearl2015}
Judea Pearl.
\newblock Detecting latent heterogeneity.
\newblock In \emph{Sociological Methods \& Research}, 2015.

\bibitem[Pearl et~al.(2016)Pearl, Glymour, and Jewell]{Pearl2016CausalII}
Judea Pearl, M~Maria Glymour, and Nicholas~P Jewell.
\newblock Causal inference in statistics: A primer.
\newblock 2016.

\bibitem[Peng et~al.(2019)Peng, Xie, Liu, Meng, Li, Yang, Yao, and Jin]{Peng2019}
Yi~Peng, Miao Xie, Jiahao Liu, Xuying Meng, Nan Li, Cheng Yang, Tao Yao, and Rong Jin.
\newblock A practical semi-parametric contextual bandit.
\newblock In \emph{Proceedings of the 28th International Joint Conference on Artificial Intelligence}, IJCAI'19, pp.\  3246–3252. AAAI Press, 2019.
\newblock ISBN 9780999241141.

\bibitem[Puli et~al.(2022)Puli, Joshi, He, and Ranganath]{Puli2022}
Aahlad Puli, Nitish Joshi, He~He, and Rajesh Ranganath.
\newblock Nuisances via negativa: Adjusting for spurious correlations via data augmentation, 2022.
\newblock URL \url{https://arxiv.org/abs/2210.01302}.

\bibitem[Radcliffe(2007)]{Radcliffe2007UsingCG}
Nicholas Radcliffe.
\newblock Using control groups to target on predicted lift: Building and assessing uplift model.
\newblock 2007.

\bibitem[Rakesh et~al.(2018)Rakesh, Guo, Moraffah, Agarwal, and Liu]{Rakesh2018}
Vineeth Rakesh, Ruocheng Guo, Raha Moraffah, Nitin Agarwal, and Huan Liu.
\newblock Linked causal variational autoencoder for inferring paired spillover effects.
\newblock In \emph{Proceedings of the 27th ACM International Conference on Information and Knowledge Management}, CIKM '18, pp.\  1679–1682, New York, NY, USA, 2018. Association for Computing Machinery.
\newblock ISBN 9781450360142.

\bibitem[Resa \& Zubizarreta(2016)Resa and Zubizarreta]{Resa2016}
María Resa and José Zubizarreta.
\newblock Evaluation of subset matching methods and forms of covariate balance.
\newblock \emph{Statistics in medicine}, 35, 07 2016.
\newblock \doi{10.1002/sim.7036}.

\bibitem[Rissanen \& Marttinen(2021)Rissanen and Marttinen]{Rissanen2021}
Severi Rissanen and Pekka Marttinen.
\newblock A critical look at the consistency of causal estimation with deep latent variable models.
\newblock 2021.

\bibitem[Robins \& Rotnitzky(1995)Robins and Rotnitzky]{Robins1995}
James Robins and A~G Rotnitzky.
\newblock Semiparametric efficiency in multivariate regression models with missing data.
\newblock \emph{Journal of The American Statistical Association - J AMER STATIST ASSN}, 90:\penalty0 122--129, 03 1995.
\newblock \doi{10.1080/01621459.1995.10476494}.

\bibitem[Robins et~al.(1994)Robins, Rotnitzky, and Zhao]{Robins1994}
James~M. Robins, Andrea Rotnitzky, and Lue~Ping Zhao.
\newblock Estimation of regression coefficients when some regressors are not always observed.
\newblock \emph{Journal of the American Statistical Association}, 89\penalty0 (427):\penalty0 846--866, 1994.
\newblock \doi{10.1080/01621459.1994.10476818}.
\newblock URL \url{https://doi.org/10.1080/01621459.1994.10476818}.

\bibitem[Robins et~al.(2000)Robins, Hern{\'a}n, and Brumback]{Robins2000}
James~M. Robins, Miguel~A. Hern{\'a}n, and Babette~A. Brumback.
\newblock Marginal structural models and causal inference in epidemiology.
\newblock \emph{Epidemiology}, 11:\penalty0 550--560, 2000.

\bibitem[Robins et~al.(2007)Robins, Sued, Lei-Gomez, and Rotnitzky]{Robins2007}
James~M. Robins, Mariela Sued, Quanhong Lei-Gomez, and Andrea Rotnitzky.
\newblock Comment: Performance of double-robust estimators when “inverse probability” weights are highly variable.
\newblock \emph{Statistical Science}, 22:\penalty0 544--559, 2007.

\bibitem[Rosenbaum \& Rubin(1983)Rosenbaum and Rubin]{ROSENBAUM1983}
Paul~R. Rosenbaum and Donald~B. Rubin.
\newblock {The central role of the propensity score in observational studies for causal effects}.
\newblock \emph{Biometrika}, 70\penalty0 (1):\penalty0 41--55, 04 1983.
\newblock ISSN 0006-3444.
\newblock \doi{10.1093/biomet/70.1.41}.

\bibitem[Rothwell(2005)]{Rothwell2005ExternalVO}
Peter~M. Rothwell.
\newblock External validity of randomised controlled trials: “to whom do the results of this trial apply?”.
\newblock \emph{The Lancet}, 365:\penalty0 82--93, 2005.

\bibitem[Rubin \& Stuart(2006)Rubin and Stuart]{Rubin2006}
Donald Rubin and Elizabeth Stuart.
\newblock Affinely invariant matching methods with discriminant mixtures of proportional ellipsoidally symmetric distributions.
\newblock \emph{The Annals of Statistics}, 34, 12 2006.
\newblock \doi{10.1214/009053606000000407}.

\bibitem[Rubin(1974)]{rubin1974estimating}
Donald~B Rubin.
\newblock Estimating causal effects of treatments in randomized and nonrandomized studies.
\newblock \emph{Journal of educational Psychology}, 66\penalty0 (5):\penalty0 688, 1974.

\bibitem[Rubin(1977)]{rubin1977assignment}
Donald~B Rubin.
\newblock Assignment to treatment group on the basis of a covariate.
\newblock \emph{Journal of educational Statistics}, 2\penalty0 (1):\penalty0 1--26, 1977.

\bibitem[Rubin(1979)]{Rubin1979UsingMM}
Donald~B. Rubin.
\newblock Using multivariate matched sampling and regression adjustment to control bias in observational studies.
\newblock \emph{Journal of the American Statistical Association}, 74:\penalty0 318--328, 1979.

\bibitem[Rubin(2005)]{rubin2005causal}
Donald~B Rubin.
\newblock Causal inference using potential outcomes: Design, modeling, decisions.
\newblock \emph{Journal of the American Statistical Association}, 100\penalty0 (469):\penalty0 322--331, 2005.

\bibitem[Rubin \& Thomas(1992)Rubin and Thomas]{RUBIN1992}
Donald~B. Rubin and Neal Thomas.
\newblock {Characterizing the effect of matching using linear propensity score methods with normal distributions}.
\newblock \emph{Biometrika}, 79\penalty0 (4):\penalty0 797--809, 12 1992.
\newblock ISSN 0006-3444.
\newblock \doi{10.1093/biomet/79.4.797}.
\newblock URL \url{https://doi.org/10.1093/biomet/79.4.797}.

\bibitem[Sachidananda \& Brunskill(2017)Sachidananda and Brunskill]{Sachidananda2017OnlineLF}
Vin Sachidananda and Emma Brunskill.
\newblock Online learning for causal bandits.
\newblock In \emph{Advances in Neural Information Processing Systems}, 2017.

\bibitem[Saito \& Yasui(2019)Saito and Yasui]{Saito2019}
Yuta Saito and Shota Yasui.
\newblock Counterfactual cross-validation: Effective causal model selection from observational data.
\newblock In \emph{arXiv preprint arXiv:1909.05299}, 2019.

\bibitem[Saturni et~al.(2014)Saturni, Bellini, Braido, Paggiaro, Sanduzzi, Scichilone, Santus, Morandi, and Papi]{Saturni2014RandomizedCT}
Sara Saturni, Federico Bellini, Fulvio Braido, Pierluigi Paggiaro, Alessandro Sanduzzi, Nicola Scichilone, Pierachille Santus, Luca Morandi, and Alberto Papi.
\newblock Randomized controlled trials and real life studies. approaches and methodologies: a clinical point of view.
\newblock \emph{Pulmonary pharmacology \& therapeutics}, 27 2:\penalty0 129--38, 2014.

\bibitem[Sawant et~al.(2018)Sawant, Namballa, Sadagopan, and Nassif]{Sawant2018}
Neela Sawant, Chitti~Babu Namballa, Narayanan Sadagopan, and Houssam Nassif.
\newblock Contextual multi-armed bandits for causal marketing.
\newblock \emph{CoRR}, abs/1810.01859, 2018.
\newblock URL \url{http://arxiv.org/abs/1810.01859}.

\bibitem[Schafer \& Kang(2009)Schafer and Kang]{Schafer2009}
Joseph Schafer and Joseph Kang.
\newblock Average causal effects from nonrandomized studies: A practical guide and simulated example.
\newblock \emph{Psychological methods}, 13:\penalty0 279--313, 01 2009.
\newblock \doi{10.1037/a0014268}.

\bibitem[SCHARFSTEIN et~al.(1999)SCHARFSTEIN, Rotnitzky, and Robins]{SCHARFSTEIN1999}
Daniel SCHARFSTEIN, A~G Rotnitzky, and James Robins.
\newblock Adjusting for nonignorable drop-out using semiparametric nonresponse models.
\newblock \emph{JASA. Journal of the American Statistical Association}, 94, 12 1999.
\newblock \doi{10.2307/2669923}.

\bibitem[Schölkopf et~al.(2021)Schölkopf, Locatello, Bauer, Ke, Kalchbrenner, Goyal, and Bengio]{Scholkopf2021}
Bernhard Schölkopf, Francesco Locatello, Stefan Bauer, Nan~Rosemary Ke, Nal Kalchbrenner, Anirudh Goyal, and Yoshua Bengio.
\newblock Towards causal representation learning, 2021.
\newblock URL \url{https://arxiv.org/abs/2102.11107}.

\bibitem[Scotina \& Gutman(2019)Scotina and Gutman]{Scotina2019}
Anthony~D. Scotina and Roee Gutman.
\newblock Matching algorithms for causal inference with multiple treatments.
\newblock \emph{arXiv preprint arXiv:1809.00269}, 2019.

\bibitem[Seaman \& Vansteelandt(2018)Seaman and Vansteelandt]{Seaman2018}
Shaun Seaman and Stijn Vansteelandt.
\newblock Introduction to double robust methods for incomplete data.
\newblock \emph{Statistical Science}, 33:\penalty0 184--197, 05 2018.
\newblock \doi{10.17863/CAM.23913}.

\bibitem[Sergey \& Christian(2015)Sergey and Christian]{Sergey2015}
Ioffe Sergey and Szegedy Christian.
\newblock Batch normalization: Accelerating deep network training by reducing internal covariate shift.
\newblock In \emph{arXiv preprint arXiv:1502.03167}, 2015.

\bibitem[Shalit et~al.(2017)Shalit, Johansson, and Sontag]{Shalit2017}
Uri Shalit, Fredrik~D. Johansson, and David Sontag.
\newblock Estimating individual treatment effect: Generalization bounds and algorithms.
\newblock ICML'17, pp.\  3076–3085. JMLR.org, 2017.

\bibitem[Shardell et~al.(2014)Shardell, Hicks, and Ferrucci]{Shardell2014}
Michelle Shardell, Gregory~E. Hicks, and Luigi Ferrucci.
\newblock {Doubly robust estimation and causal inference in longitudinal studies with dropout and truncation by death}.
\newblock \emph{Biostatistics}, 16\penalty0 (1):\penalty0 155--168, 07 2014.
\newblock \doi{10.1093/biostatistics/kxu032}.
\newblock URL \url{https://doi.org/10.1093/biostatistics/kxu032}.

\bibitem[Shimodaira(2000)]{Shimodaira2000}
Hidetoshi Shimodaira.
\newblock Improving predictive inference under covariate shift by weighting the log-likelihood function.
\newblock \emph{Journal of Statistical Planning and Inference}, 90:\penalty0 227--244, 2000.

\bibitem[Sibbald \& Roland(1998)Sibbald and Roland]{Sibbald1998UnderstandingCT}
Bonnie Sibbald and Martin Roland.
\newblock Understanding controlled trials: Why are randomised controlled trials important?
\newblock \emph{BMJ}, 316:\penalty0 201, 1998.

\bibitem[Simon(2001)]{Simon2001IsTR}
Stephen~D. Simon.
\newblock Is the randomized clinical trial the gold standard of research?
\newblock \emph{Journal of andrology}, 22 6:\penalty0 938--43, 2001.

\bibitem[Slivkins(2019)]{Slivkins2019}
Aleksandrs Slivkins.
\newblock Introduction to multi-armed bandits.
\newblock \emph{CoRR}, abs/1904.07272, 2019.
\newblock URL \url{http://arxiv.org/abs/1904.07272}.

\bibitem[Stefanski \& Boos(2002)Stefanski and Boos]{Stefanski2002}
Leonard~A. Stefanski and Dennis~D. Boos.
\newblock The calculus of m-estimation.
\newblock \emph{The American Statistician}, 56:\penalty0 29 -- 38, 2002.

\bibitem[Stewart \& Parmar(1996)Stewart and Parmar]{Stewart1996BiasIT}
Lesley~A Stewart and Mahesh K.~B. Parmar.
\newblock Bias in the analysis and reporting of randomized controlled trials.
\newblock \emph{International Journal of Technology Assessment in Health Care}, 12:\penalty0 264 -- 275, 1996.

\bibitem[Stojanov et~al.(2021)Stojanov, Li, Gong, Cai, Carbonell, and Zhang]{Stojanov2021}
Petar Stojanov, Zijian Li, Mingming Gong, Ruichu Cai, Jaime Carbonell, and Kun Zhang.
\newblock Domain adaptation with invariant representation learning: What transformations to learn?
\newblock In M.~Ranzato, A.~Beygelzimer, Y.~Dauphin, P.S. Liang, and J.~Wortman Vaughan (eds.), \emph{Advances in Neural Information Processing Systems}, volume~34, pp.\  24791--24803. Curran Associates, Inc., 2021.
\newblock URL \url{https://proceedings.neurips.cc/paper/2021/file/cfc5d9422f0c8f8ad796711102dbe32b-Paper.pdf}.

\bibitem[Stuart(2010)]{Stuart2010}
Elizabeth~A. Stuart.
\newblock {Matching Methods for Causal Inference: A Review and a Look Forward}.
\newblock \emph{Statistical Science}, 25\penalty0 (1):\penalty0 1 -- 21, 2010.
\newblock \doi{10.1214/09-STS313}.
\newblock URL \url{https://doi.org/10.1214/09-STS313}.

\bibitem[Swaminathan \& Joachims(2015)Swaminathan and Joachims]{Swaminathan2015}
Adith Swaminathan and Thorsten Joachims.
\newblock Counterfactual risk minimization: Learning from logged bandit feedback.
\newblock \emph{CoRR}, abs/1502.02362, 2015.
\newblock URL \url{http://arxiv.org/abs/1502.02362}.

\bibitem[Swaminathan et~al.(2017)Swaminathan, Krishnamurthy, Agarwal, Dud\'{\i}k, Langford, Jose, and Zitouni]{Swaminathan2017}
Adith Swaminathan, Akshay Krishnamurthy, Alekh Agarwal, Miroslav Dud\'{\i}k, John Langford, Damien Jose, and Imed Zitouni.
\newblock Off-policy evaluation for slate recommendation.
\newblock NIPS'17, pp.\  3635–3645, Red Hook, NY, USA, 2017. Curran Associates Inc.
\newblock ISBN 9781510860964.

\bibitem[Tan(2010)]{Tan2010BoundedEA}
Zhiqiang Tan.
\newblock Bounded, efficient and doubly robust estimation with inverse weighting.
\newblock \emph{Biometrika}, 97:\penalty0 661--682, 2010.

\bibitem[Trifunov et~al.(2020)Trifunov, Shadaydeh, Runge, Eyring, Reichstein, and Denzler]{Trifunov2020}
Violeta~Teodora Trifunov, Maha Shadaydeh, Jakob Runge, Veronika Eyring, Markus Reichstein, and Joachim Denzler.
\newblock Causal link estimation under hidden confounding in ecological time series.
\newblock 01 2020.

\bibitem[Trifunov et~al.(2022)Trifunov, Shadaydeh, and Denzler]{Trifunov2022}
Violeta~Teodora Trifunov, Maha Shadaydeh, and Joachim Denzler.
\newblock Time series causal link estimation under hidden confounding using knockoff interventions, 2022.

\bibitem[Uri et~al.(2017)Uri, Fredrik~D., and Songtag]{Uri2017}
Shalit Uri, Johansson Fredrik~D., and David Songtag.
\newblock Estimating individual treatment effect: generalization bounds and algorithms.
\newblock In \emph{arXiv preprint arXiv:1606.03976}, 2017.

\bibitem[van Amsterdam et~al.(2022)van Amsterdam, de~Jong, Verhoeff, Leiner, and Ranganath]{Amsterdam2022}
Wouter A.~C. van Amsterdam, Pim~A. de~Jong, Joost J.~C. Verhoeff, Tim Leiner, and Rajesh Ranganath.
\newblock Decision making in cancer: Causal questions require causal answers, 2022.
\newblock URL \url{https://arxiv.org/abs/2209.07397}.

\bibitem[Veitch et~al.(2020)Veitch, Sridhar, and Blei]{Veitch2020a}
Victor Veitch, Dhanya Sridhar, and David Blei.
\newblock Adapting text embeddings for causal inference.
\newblock In Jonas Peters and David Sontag (eds.), \emph{Proceedings of the 36th Conference on Uncertainty in Artificial Intelligence (UAI)}, volume 124 of \emph{Proceedings of Machine Learning Research}, pp.\  919--928. PMLR, 03--06 Aug 2020.

\bibitem[Villani(2009)]{villani2009optimal}
C{\'e}dric Villani.
\newblock \emph{Optimal transport: old and new}, volume 338.
\newblock Springer, 2009.

\bibitem[Vowels et~al.(2020)Vowels, Camg{\"o}z, and Bowden]{Vowels2020}
Matthew~James Vowels, Necati~Cihan Camg{\"o}z, and R.~Bowden.
\newblock Targeted vae: Structured inference and targeted learning for causal parameter estimation.
\newblock \emph{ArXiv}, abs/2009.13472, 2020.

\bibitem[Waernbaum \& Pazzagli(2017)Waernbaum and Pazzagli]{Waernbaum2017ModelMA}
Ingeborg Waernbaum and Laura Pazzagli.
\newblock Model misspecification and bias for inverse probability weighting and doubly robust estimators.
\newblock \emph{arXiv: Statistics Theory}, 2017.

\bibitem[Wang et~al.(2016)Wang, Zhi, and Eric~Tchetgen]{Wang2016}
Miao Wang, Geng Zhi, and Tchetgen Eric~Tchetgen.
\newblock Identifying causal effects with proxy variables of an unmeasured confounder.
\newblock In \emph{arXiv preprint arXiv:1705.08821}, 2016.

\bibitem[Wang \& Jordan(2021)Wang and Jordan]{Wang2021}
Yixin Wang and Michael~I. Jordan.
\newblock Desiderata for representation learning: A causal perspective, 2021.

\bibitem[Weitzen et~al.(2004)Weitzen, Lapane, Toledano, Hume, and Mor]{Weitzen2004}
Sherry Weitzen, Kate Lapane, Alicia Toledano, Anne Hume, and Vincent Mor.
\newblock Principles for modeling propensity scores in medical research: A systematic literature review.
\newblock \emph{Pharmacoepidemiology and drug safety}, 13:\penalty0 841--53, 12 2004.
\newblock \doi{10.1002/pds.969}.

\bibitem[Wu et~al.(2020)Wu, Kuang, Yuan, Li, Zhou, Tao, Zhu, Zhuang, and Wu]{Wu2020}
Anpeng Wu, Kun Kuang, Junkun Yuan, Bo~Li, Pan Zhou, Jianrong Tao, Qiang Zhu, Yueting Zhuang, and Fei Wu.
\newblock Learning decomposed representation for counterfactual inference.
\newblock \emph{ArXiv}, abs/2006.07040, 2020.

\bibitem[Wu \& Fukumizu(2021)Wu and Fukumizu]{Wu2021IntactVAEET}
Pengzhou~(Abel) Wu and Kenji Fukumizu.
\newblock Intact-vae: Estimating treatment effects under unobserved confounding.
\newblock \emph{ArXiv}, abs/2101.06662, 2021.

\bibitem[Yao et~al.(2018)Yao, Li, Li, Huai, Gao, and Zhang]{Yao2018}
Liuyi Yao, Sheng Li, Yaliang Li, Mengdi Huai, Jing Gao, and Aidong Zhang.
\newblock Representation learning for treatment effect estimation from observational data.
\newblock In S.~Bengio, H.~Wallach, H.~Larochelle, K.~Grauman, N.~Cesa-Bianchi, and R.~Garnett (eds.), \emph{Advances in Neural Information Processing Systems}, volume~31. Curran Associates, Inc., 2018.
\newblock URL \url{https://proceedings.neurips.cc/paper/2018/file/a50abba8132a77191791390c3eb19fe7-Paper.pdf}.

\bibitem[Yao et~al.(2020)Yao, Alexis, and Mihaela~van der]{Zhang2020}
Zhang Yao, Bellot Alexis, and Schaar Mihaela~van der.
\newblock Learning overlapping representations for the estimation of individualized treatment effects.
\newblock In \emph{International Conference of Machine Learning}, 2020.

\bibitem[Yazdani \& Boerwinkle(2015)Yazdani and Boerwinkle]{Yazdani2015}
A~Yazdani and E~Boerwinkle.
\newblock Causal inference in the age of decision medicine.
\newblock \emph{Journal of data mining in genomics \& proteomics}, 6, 2015.

\bibitem[Ye et~al.(2020)Ye, Lin, Xie, and Lui]{Ye2020}
Li~Ye, Yishi Lin, Hong Xie, and John~C.s Lui.
\newblock Combining offline causal inference and online bandit learning for data driven decisions.
\newblock 01 2020.
\newblock URL \url{http://arxiv.org/abs/2001.05699}.

\bibitem[Yoon et~al.(2018)Yoon, Jordon, and van~der Schaar]{Yoon2018}
Jinsung Yoon, James Jordon, and Mihaela van~der Schaar.
\newblock Ganite: Estimation of individualized treatment effects using generative adversarial nets.
\newblock In \emph{International Conference on Learning Representations}, 2018.

\bibitem[Zeng et~al.(2020)Zeng, Assaad, Tao, Datta, Carin, and Li]{Zeng2020DoubleRR}
Shuxi Zeng, Serge Assaad, Chenyang Tao, Shounak Datta, Lawrence Carin, and Fan Li.
\newblock Double robust representation learning for counterfactual prediction.
\newblock \emph{ArXiv}, abs/2010.07866, 2020.

\bibitem[Zhang et~al.(2020)Zhang, Kumor, and Bareinboim]{Zhang2020b}
Junzhe Zhang, Daniel Kumor, and Elias Bareinboim.
\newblock Causal imitation learning with unobserved confounders.
\newblock In \emph{Proceedings of the 34th International Conference on Neural Information Processing Systems}, 2020.

\bibitem[Zhang et~al.(2021)Zhang, Berrevoets, and van~der Schaar]{Zhang2021IdentifiableER}
Yao Zhang, Jeroen Berrevoets, and Mihaela van~der Schaar.
\newblock Identifiable energy-based representations: An application to estimating heterogeneous causal effects.
\newblock \emph{ArXiv}, abs/2108.03039, 2021.

\bibitem[Zhao(2004)]{Zhao2004}
Zhong Zhao.
\newblock Using matching to estimate treatment effects: Data requirements, matching metrics, and monte carlo evidence.
\newblock \emph{The Review of Economics and Statistics}, 86:\penalty0 91--107, 02 2004.
\newblock \doi{10.1162/003465304323023705}.

\bibitem[Zheng \& Kleinberg(2019)Zheng and Kleinberg]{Zheng2019}
Min Zheng and Samantha Kleinberg.
\newblock Using domain knowledge to overcome latent variables in causal inference from time series.
\newblock In Finale Doshi-Velez, Jim Fackler, Ken Jung, David Kale, Rajesh Ranganath, Byron Wallace, and Jenna Wiens (eds.), \emph{Proceedings of the 4th Machine Learning for Healthcare Conference}, volume 106 of \emph{Proceedings of Machine Learning Research}, pp.\  474--489. PMLR, 09--10 Aug 2019.
\newblock URL \url{https://proceedings.mlr.press/v106/zheng19a.html}.

\bibitem[Zheng et~al.(2020)Zheng, Marsh, Nickerson, and Kleinberg]{Zheng2020HowCI}
Min Zheng, Jessecae~K. Marsh, Jeffrey~V. Nickerson, and Samantha Kleinberg.
\newblock How causal information affects decisions.
\newblock \emph{Cognitive Research: Principles and Implications}, 5, 2020.

\bibitem[Zhihong \& Manabu(2012)Zhihong and Manabu]{Cai2012}
Cai Zhihong and Kuroki Manabu.
\newblock On identifying total effects in the presence of latent variables and selection bias.
\newblock In \emph{arXiv preprint arXiv:1206.3239}, 2012.

\bibitem[Zhong et~al.(2022)Zhong, Xiao, Ren, Liang, Yao, Yang, and Cen]{Zhong2022DESCNDE}
Kailiang Zhong, Fengtong Xiao, Yan Ren, Yaorong Liang, Wenqing Yao, Xiaofeng Yang, and Ling Cen.
\newblock Descn: Deep entire space cross networks for individual treatment effect estimation.
\newblock \emph{Proceedings of the 28th ACM SIGKDD Conference on Knowledge Discovery and Data Mining}, 2022.

\bibitem[Zubizarreta(2012)]{Zubizarreta2012}
José Zubizarreta.
\newblock Using mixed integer programming for matching in an observational study of kidney failure after surgery.
\newblock \emph{JASA. Journal of the American Statistical Association}, 107, 12 2012.
\newblock \doi{10.1080/01621459.2012.703874}.

\end{thebibliography}
\bibliographystyle{tmlr}
\appendix

\section{Appendix}
\subsection{Naive ATE estimation}
\label{chp1:agg}
Having an RCT dataset, by default gives us
\begin{align*}
    \lbrace Y(0), Y(1)\rbrace \indep A | X=x \text{, for all } x\in \mathcal{X}.
\end{align*}
We can compute ATE by aggregating differences in mean estimators of treatment and control group
\begin{align*}
    ATE &= \mathbb{E}_{p(x)} [\mathbb{E}[ Y(1) -Y(0) |X=x]]\\
        &= \mathbb{E}_{p(x)} \left[\mathbb{E}[Y(1)|X=x] - \mathbb{E}[Y(0) |X=x]\right]
\end{align*}
Taking the Monte Carlos approximation, we have
\begin{align*}
    \widehat{ATE} =&  \mathbb{E}_p(x) \left[\frac{1}{N_{x1}} \sum_{i:t_i=1} y_i -  \frac{1}{N_{x0}} \sum_{j:t_j=0} y_j\right]\\
    =&  \frac{N_x}{N} \sum_x \left[\frac{1}{N_{x1}} \sum_{i:t_i=1} y_i -  \frac{1}{N_{x0}} \sum_{j:t_j=0} y_j\right]\\
\end{align*}
where $N_{xt}$ is the number of $x$ data points with treatment assignment $t$, $N_x$ is number of $x$ data points, $N$ is the total data points.
For simplicity, let $\bar{e}(x) = N_{x1}$ and $1-\bar{e}(x) = N_{x0}$.
We can re-express the $\widehat{ATE}(x)$ as 
\begin{align*}
    \widehat{ATE}_\text{AGG}(x) = \frac{1}{\bar{e}(x)} \sum_{i:t_i=1} y_i -  \frac{1}{1-\bar(e)(x)} \sum_{j:t_j=0} y_j.
\end{align*}
Our $\widehat{ATE}(x)$ is unbiased estimator and the estimation error is
\begin{align*}
    \sqrt{N_x} (\widehat{ATE}_\text{AGG}(x) - ATE(x)) \rightarrow \mathcal{N}\left(0,
    \frac{\text{Var}[Y(1)|X=x]}{\bar{e}(x)} + \frac{\text{Var}[Y(0)|X=x]}{1-\bar{e}(x)}\right).
\end{align*}
Under the assumption that $\text{Var}[Y(A)|X=x]=\sigma^2(x)$ does not 
depend on $A$, then we
\begin{align*}
    \sqrt{N_x} (\widehat{ATE}_\text{AGG}(x) - ATE(x)) \rightarrow \mathcal{N}\left(0,
    \frac{\sigma^2(x)}{\bar{e}(x)(1-\bar{e}(x))}\right).
\end{align*}

We can re-write our estimator by decomposing in terms of true ATE and the approximation error:
\begin{align*}
    \widehat{ATE}_\text{AGG} &= \sum_x \frac{N_x}{N} \widehat{ATE}_{\text{AGG}}(x)\\
        &= \sum_x p(x)ATE(x) + 
        \underbrace{\sum_x p(x)(ATE(x) - \widehat{ATE}(x))}_{\approx \mathcal{N}(0,\sum_x p(x)^2\text{Var}[\widehat{ATE}(x)]} \\
         & \qquad + \underbrace{ \sum_x \left(p(x) - \frac{N_x}{N}\right)ATE(x)}_{\approx \mathcal{N}(0, N^{-1} \text{Var}[ATE(x)]} 
         + \underbrace{\sum_x \left(p(x) - \frac{N_x}{N}\right)(ATE(x) - \widehat{ATE}(x))}_{=\mathcal{O}(N^{-1})}.
\end{align*}    
This makes $\widehat{ATE}_\text{AGG}$ error to distribute $\mathcal{N}(0,\text{Var}_{AGG})$, where 
\begin{align}
    N\text{Var}_{AGG} &= \text{Var}[ATE_{\text{AGG}}(x)] + \mathbb{E}_p(x)\left[ \frac{\sigma^2(x)}{\bar{e}(x)(1-\bar{e}(x))}\right].
    \label{eqn:chp1:var_agg}
\end{align}

\subsection{ATE ordinary least-squares estimator}

Rather than taking the difference in direct estimations of expected potential outcomes, 
we will model the potential outcome using linear regression,
\begin{align*}
    Y_i(t) = c_t + X_i\beta_t + \epsilon_i(t)
\end{align*}
where $\mathbb{E}[\epsilon_i(t)|X_i]=0$ and $\text{Var}[\epsilon_i(t)|X_i]=\sigma^2$.
The mean of data to be zero $\mathbb{E}[X]=0$ by normalizing the dataset and the variance of data is $\text{Var}[X]=\sigma_X$.
Because we are in RCT setting, $p(T=1) = p(T=0)=\frac{1}{2}$.

In this setup, we can write the ATE as
\begin{align*}
    ATE = \mathbb{E}[Y(1)-Y(0)] = c_1 - c_0 + \mathbb{E}[X](\beta_1 - \beta_0).
\end{align*}
Now we can run ordinary least-square (OLS) estimator to estimate $c_t$ and $\beta_t$,
\begin{align*}
    \hat{\tau}_{OLS} = \hat{c}_1 -\hat{c}_0 + \bar{X}(\hat{\beta}_1 - \hat{\beta}_0),
\end{align*}
where $\bar{X} =\frac{1}{n}\sum^{n}_{i=1} x_i$.
The standard error for $c_t$ and $\beta_t$ estimation are
\begin{align*}
    \hat{c_1}-c_1 = \mathcal{N}\left(0,\frac{\sigma^2}{n_1}\right) \text{ and }
    \hat{c_0}-c_0 = \mathcal{N}\left(0,\frac{\sigma^2}{n_0}\right)
\end{align*}
respectively. 
Since $\mathbb{E}[X]=0$ and $\bar{X}$ asymptotically approaches to 0 with the standard error of $\|\beta_1 \beta_0\|^2_{\sigma_X} / N$, we have
\begin{align*}
    \hat{\tau}_{OLS} - \tau = \hat{c_1}-c_1 - \hat{c_0}-c_0 + \bar{X}(\beta_1 -\beta_0) + \bar{X}(\hat{\beta}_1 -\beta_1 - \hat{\beta}_0 + \beta_0)
\end{align*}
where the last term has $O(1/n)$ error rate.
This makes $\widehat{ATE}$ error to distribute $\mathcal{N}(0, \text{Var}_{OLS})$, where $\text{Var}_{OLS})=4\sigma^2 + \|\beta_0-\beta_1\|_{\sigma_X}^2$.

\subsection{Active CI Learning}
\subsubsection{Contextual Bandit Algorithm}
\label{app:bandits_algo}

Contextual bandit is known as an online method that helps you make decisions in given contexts. 
It finds optimal decisions by maximizing their total rewards over a period of time.
In contextual bandit setup, we get an observation $x_t$ at time $t$, the algorithm picks an action from a finite set $a_t \in \mathcal{A}$, and then executes action $a_t$ on observation $x_t$. 
The reward $y_t \in [0,1]$ is given by the world which is some distribution parameterized by $(X, A)$ variables and the samples are drawn independently and identically. 
The reward depends on both the context $x_t$ and the chosen action $a_t$ at each round. The reward distribution can change over time but this change is explained by the stream of context data\footnote{The reward and outcome can be viewed the same and used interchangeably.}. 
The action is chosen based on the choice of algorithm, such as UBC1 and Thomson sampling methods \cite{Agrawal2013}\footnote{We will not review the details of each algorithm in this paper but refer the reader to \cite{Bouneffouf2020}.} (see Appendix~\ref{app:bandits_algo}).

{\bf Examples of contextual bandits' policy algorithms}\\

\begin{algorithm}[h!]
\caption{Epsilon-Greedy}\label{alg:epsilon_greedy_ci}
\begin{algorithmic}
        \State Toss a coin with success probability $\epsilon_t$.
        \If {success}
        \State explore: choose an arm uniformly at random.
        \Else
        \State exploit: choose the arm with the highest average reward so far.
        \EndIf
\end{algorithmic}
\end{algorithm}

\begin{algorithm}[h!]
\caption{“High-confidence elimination”}\label{alg:hce_ci}
\begin{algorithmic}
        \State Alternate two arms until $UCB(a_t) < LCB(a_t^\prime)$ after some even round t.
        \State Abandon arm $a$, and use arm $a^\prime$ forever since.
\end{algorithmic}
\end{algorithm}

\begin{algorithm}[h!]
\caption{UBC1}\label{alg:ubc1_ci}
\begin{algorithmic}
        \State pick arm some a which maximizes $UCB(a_t)$
\end{algorithmic}
\end{algorithm}

\begin{algorithm}[h!]
\caption{Thompson Sampling}\label{alg:thompson_sampling_ci}
\begin{algorithmic}
        \State Sample mean reward vector $\mathbb{E}[Y(a)]$ for each action $a$ from the posterior distribution $p(a|X)$.
        \State Choose the best arm $a_t$ according to $\mathbb{E}[Y(a_t)]$.
\end{algorithmic}
\end{algorithm}

{\bf Upper bound on ATE}\\

Both {\it high-confidence elimination method} and {\it UCB} algorithms have assured to be upper-bounded for ATE estimations since the potential outcomes are distanced enough that the two confidence intervals do not overlap,
\begin{align*}
    ATE = \left|\mathbb{E}\left[ Y(A=a) \right]-\mathbb{E}\left[ Y(A=a^\prime) \right] \right| \leq 2(r_t(a) + r_t(a^\prime)) \leq 4(\sqrt{2\log(T)/\lfloor t/2 \rfloor}) \leq O(\sqrt{\log(T)/t}).
\end{align*}
where $T$ is the total iterations and $n_t(a) = \lfloor \frac{t}{2} \rfloor$ since $a$ and $a^\prime$ has been altered.
In this case, we can aggregate data into an intervention dataset until the confidence interval does not overlap and use them to estimate ATE.

\subsection{Passive CI Learning}
\subsubsection{Inverse Probability Weighting}
\label{app:ipw}
    Since propensity score $e(X)$ is unknown in practice, we have to estimate $\hat{e}(X)$ via parametric and non-parametric regression.
    We use non-parametric regression to estimate $\hat{e}(X) = \frac{N_1}{N}$ where $N_1=\sum^{i=1}\mathbb{I}[Y(X)=1]$.
    ATE estimation is
        \begin{align*}
            \widehat{ATE}_{\text{IPW}} =  \frac{1}{N}\sum_{i}^{n}\left[\frac{\mathbb{I}[A_i=1]y_i(1)}{\hat{e}(x_i)}\right] - \frac{1}{N}\sum_{i}^{n} \left[\frac{\mathbb{I}[A_i=0]y_i(0)}{1-\hat{e}(x_i)}\right].
        \end{align*}

   Suppose that $\hat{e}(x) \rightarrow e(x)$ as $N \rightarrow \infty$ (i.e., $\sup_{x \in \mathcal{X}}|e(x) - \hat{e(x)}| \rightarrow \mathcal{O}(a_n)$. Then, the ATE error becomes
        \begin{align*}
            |ATE - \widehat{ATE}_{\text{IPW}}| = \mathcal{O}\Bigg(\frac{a_n}{\eta}\Bigg)
        \end{align*}
        where $\eta \leq e(x) \leq 1-\eta$ for all $x\in\mathcal{X}$
        and $|Y_i| \leq 1$.
        Therefore, $\widehat{ATE}$ is concentrated at $\frac{1}{\sqrt{n}}$.

        The weighted population outcomes of two actions, $A=0$ and $A=1$, is an unbiased estimate of the average treatment effect
        \begin{align}
            ATE = \mathbb{E}\left[\frac{\mathbb{I}[A=1]Y(1)}{e(X)}\right] - \mathbb{E}\left[\frac{\mathbb{I}[A=0]Y(0)}{1-e(X)}\right].
            \label{eq:ATE_IPW}
        \end{align}
        In order to express the variance of $ATE$ in Equation~\ref{eq:ATE_IPW}, let us re-express $Y(0)$ and $Y(1)$,
        \begin{align*}
            Y(0) &= c(X) - (1-e(X))ATE(X) + \epsilon(0)\\
            Y(1) &= c(X) + e(X)ATE(X) + \epsilon(1)
        \end{align*}
        where $c(X)$ is a function that makes the expression above work, and
        $\mathbb{E}[\epsilon(0)|X]=0$ and $\mathbb{E}[\epsilon(1)|X]=0$. assume that $\text{Var}(\epsilon(A)|X)=\sigma^2(X)$ does not depend on $A$.
        Then 
        \begin{align*}
            N \text{Var}_{IPW}[ATE(X)] =& \text{Var}\left[\frac{AX}{e(X)} - \frac{(1-A)Y}{1-e(X)}\right]\\
                            =& \text{Var}\left[\frac{Ac(X)}{e(X)} - \frac{(1-A)c(X)}{1-e(X)}\right]
                            + \text{Var}[ATE(X)] + \text{Var}\left[\frac{A\epsilon(X)}{e(X)} - \frac{(1-A)\epsilon(X)}{1-e(X)}\right]\\
                            =& \mathbb{E}\left[\frac{c(X)^2}{e(X)(1-e(X))}\right] + \text{Var}[ATE(X)] +\mathbb{E}\left[\frac{\sigma^2(X)}{e(X)(1-e(X)}\right]
        \end{align*}
        Note that we can express the variance of IPW estimator  in terms of 
        the variance of our estimator is worse than the variance
        of aggregating difference in mean estimator $\text{Var}_{AGG}$
        in Equation~\ref{eqn:chp1:var_agg}, which is the naive ATE estimator
        for RCT dataset,
        \begin{align*}
            N \text{Var}_{IPW}[ATE(X)] = N \text{Var}_{AGG}[ATE(X)] + \mathbb{E}\left[\frac{c(X)^2}{e(X)(1-e(X))}\right]. 
        \end{align*}
        This means that IPW has higher variance than AGG estimator.
        Surprisingly, we can conclude that the true propensity score performs worse than empirical propensity score, since AGG estimator used $\bar{e}(x)=\frac{N_x}{N}$ (see Section~\ref{chp1:agg}).

\subsubsection{Domain Invariance Regularization}
\label{app:domain_invariance_reg}

Intuitively, inducing the treated and control representational distribution to be the same is that it induces the two learned prediction function $p_\theta(y|t=0,x)$ and $p_\theta(y|t=1,x)$ to have better generalization across the treated and control populations. Indeed, \cite{Shalit2017} show that CFR objective function is the upper bounds of the PEHE generalization error \cite{Bareinboim2016},
\begin{align*}
    PEHE \leq 2(\mathcal{L}_{\text{F}} + \mathcal{L}_{\text{CF}} - 2\sigma^2_Y)
    \leq 2(\mathcal{L}_{\text{F}}^{T=0} + \mathcal{L}_{\text{F}}^{T=1} + B_h \text{IPM}_\mathcal{F} (\hat{p}_{T=1}(h(X)), \hat{p}_{T=0}(h(X)) 
\end{align*}
where the expected factual and counterfactual losses are defined as
\begin{align*}
    \mathcal{L}_\text{F} &= \mathbb{E}_{p(X,T)}[ l(X,T) ] = u \mathcal{L}_\text{F}^{T=1} + (1-u)\mathcal{L}_\text{F}^{T=0}\\
    \mathcal{L}_\text{CF} &= \mathbb{E}_{p(X,1-T)}[ l(X,T) ] = (1-u) \mathcal{L}_\text{CF}^{T=0} + u \mathcal{L}_\text{CF}^{T=1}
\end{align*}
respectively. The expected loss function for individual data point over $p(Y_T|X=x)$ is $l(X=x,T=t)$ and $u=p(T=1)$ is the proportions of treated in the population. 
The expected factual/counterfactual treated and control losses becomes
\begin{align*}
  \mathcal{L}_\text{F}^{T=1} &= \mathbb{E}_{p(X,1)}[ l(X,1) ],  \qquad
  \mathcal{L}_\text{F}^{T=0} = \mathbb{E}_{p(X,0)}[ l(X,0) ] \\
  \mathcal{L}_\text{CF}^{T=1} &= \mathbb{E}_{p(X,0)}[ l(X,1) ], \qquad
  \mathcal{L}_\text{CF}^{T=0} = \mathbb{E}_{p(X,1)}[ l(X,0) ] 
\end{align*}
 respectively.
$\sigma^2_Y := \text{min}\lbrace \sigma^2_0, \sigma^2_1 \rbrace$ and $\sigma^2_t = \mathbb{E}_{p(X,T)}[(Y-f_\phi(X))^2]$ is the expected variance of $Y_T$. 
The full proof can be found in the original paper \cite{Shalit2017} but the key idea is that $\mathcal{L}_\text{CF} \leq  u \mathcal{L}_\text{F}^{T=0} + (1-u)\mathcal{L}_\text{F}^{T=1} + B_h \text{IPM}_\mathcal{F}$.

\subsection{Posterior Variance Reduction Regularization}
\label{app:variance_reduction_reg}

\begin{figure}[t]
\begin{minipage}{\textwidth}
\begin{center}
    \includegraphics[width=0.6\textwidth]{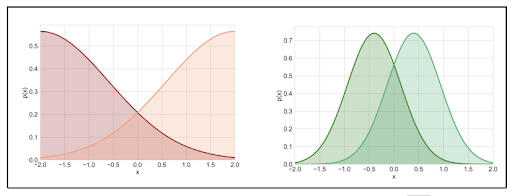}
  \end{center}
  \caption{Red distribution (left) has large overlap on the tails and the green distribution has small overlap on the tails }
\label{chp2:fig:toy_dists}
\end{minipage}\hfill

\begin{minipage}{0.49\textwidth}
\centering
\includegraphics[width=\textwidth]{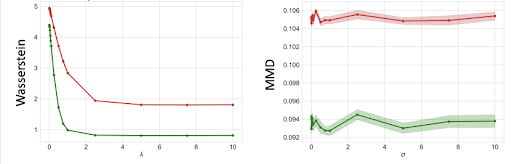}
(a)\label{fig:compare-ddsm-m}
\end{minipage}\hfill
\begin{minipage}{0.49\textwidth}
\centering
\includegraphics[width = \textwidth]{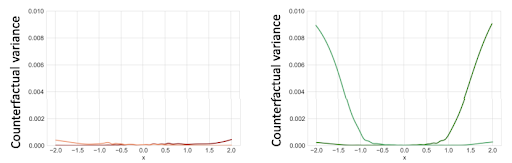}
(b)
\end{minipage}
\caption{(a) Results using IPMs - Wasserstein distance and MMD, (b) Results using counterfactual variance }
\label{chp2:fig:toy_exp}
\end{figure}

\cite{Zhang2020} demonstrate why distributional distance to balance the treated and controlled representations is not ideal using an toy example in Figure~\ref{chp2:fig:toy_dists}.
The red population comes from two truncated normal distributions having large overlap in tails and the green population comes from two normal distributions having small overlap in the tails. 
Figure \ref{chp2:fig:toy_exp}(a) illustrates that both the MMD \cite{Gretton12a} and Wasserstein distances \cite{villani2009optimal} are smaller in the green population compared to the red population, even though sufficient support is satisfied in the red population and not the green population.
In contrast, counterfactual variance perfectly describes the lack of support in the red population as shown in Figure \ref{chp2:fig:toy_exp}(b).

Minimizing the counterfactual variance can lead to better generalization error \cite{Zhang2020}. 
The follow Theorem shows that the counterfactual Gibbs risk $R_{p(1-T)}$ is upper bounded by two terms that corresponds to domain invariance and counterfactual variance,
\begin{align*}
    R_{p(1-t)} \leq \sup_x \frac{p(x,1-t)}{p(x,t)} \mathcal{L}_{\text{Factual}} + \frac{1}{2}\mathbb{E}_{p(x)}[\sigma^2(x|\mathcal{X}, \theta)].
\end{align*}
We observe that the first term consists of the factual loss and distribution mismatch. Minimizing both factual loss and the making posterior distribution to be invariant will lead to lower counterfactual Gibbs risk.
The second term corresponds to counterfactual variance. This illustrates that minimizing the counterfactual variance is indispensable regularization term as well. \\

\subsection{Deep Latent-Variable Model: UTVAE}
\label{app:utvae}
In the CEVAE, there are two conditional distributions that depend on treatment $T$, $p_\theta(Y|T,Z)$ and $q_\phi(Z|T,X,Y)$. 
Both of these distributions can be estimated using samples drawn from a treatment distribution that is either dependent on or independent of the confounding factor.
In doing so, we have the option to use observational data based, or uniform treatment distributions, for estimating generative and inference distributions respectively
\begin{align*}
    \mathcal{L}(\theta; \phi) &= \mathcal{L}_{\text{CEVAE}}(\theta; \bar{\phi}) + \mathcal{L}_{\text{UTVAE}}(\phi; \bar{\theta})
\end{align*}
where $\bar{\theta}$ and $\bar{\phi}$ are fixed parameters - the gradients  with respect to these variables are blocked in the computational graph.
We do so in order to isolate the impact of the choice of treatment distribution on the associated conditional distributions.

\end{document}